\documentclass[journal]{IEEEtran}

\usepackage{cite}
\usepackage[utf8]{inputenc}
\usepackage{blkarray}
\usepackage{lastpage}
\usepackage{ifthen}
\usepackage{caption}
\usepackage{float}
\usepackage{amsmath}
\usepackage{ wasysym }
\usepackage{ amssymb }
\usepackage{ upgreek }
\usepackage[linesnumbered]{algorithm2e}
\usepackage{bbm}
\usepackage{algpseudocode}
\usepackage{pifont}
\usepackage{enumerate}
\usepackage{tikz}
\usetikzlibrary{automata,arrows,positioning,calc}
\usetikzlibrary{arrows,matrix,positioning}
\usepackage{wrapfig}
\usepackage{cancel}
\usepackage{pifont} 
\usepackage{subcaption}
\usepackage{hyperref}
\usepackage{pgfplots}
\usepackage{cancel}
\usepackage{color}
\usepackage{mathrsfs}

\newcommand{\bpg}{\begin{paragraph}{}}
\newcommand{\epg}{\end{paragraph}}
\newcommand{\setpg}[1]{\begin{paragraph}{\bf Question #1:}}
\newcommand{\unsetpg}{\end{paragraph}\newpage}
\newcommand{\bit}{\begin{itemize}}
\newcommand{\eit}{\end{itemize}}
\newcommand{\beq}{\begin{equation}}
\newcommand{\eeq}{\end{equation}}
\newcommand{\beqn}{\begin{equation*}}
\newcommand{\eeqn}{\end{equation*}}
\newcommand{\beqa}{\begin{equation}\begin{aligned}}
\newcommand{\eeqa}{\end{aligned}\end{equation}}
\newcommand{\beqna}{\begin{equation*}\begin{aligned}}
\newcommand{\eeqna}{\end{aligned}\end{equation*}}
\newcommand{\bal}{\begin{align}}
\newcommand{\eal}{\end{align}}
\newcommand{\baln}{\begin{align*}}
\newcommand{\ealn}{\end{align*}}
\newcommand{\bether}{\begin{gather*}}
\newcommand{\eether}{\end{gather*}}

\newcommand{\ii}{\item}

\newcommand{\enum}{\begin{enumerate}}
\newcommand{\enuma}{\begin{enumerate}[(a)]}
\newcommand{\eenum}{\end{enumerate}}
\newcommand{\norm}[1]{||#1||}

\DeclareMathOperator{\classin}{\textbf{class}^k_{\text{in}}}
\DeclareMathOperator{\classout}{\textbf{class}^k_{\text{out}}}
\newcommand{\lp}{\left(}
\newcommand{\rp}{\right)}

\newcommand{\matrixcolsep}[1]{\kern#1em\vline}

\newcommand{\smallb}{{\scriptscriptstyle|}}

\newcommand{\reals}{\mathbb{R}}

\newcommand{\dual}[1]{\widetilde{#1}}

\newcommand{\bolx}{\boldsymbol{x}}

\newcommand{\bolb}{\boldsymbol{\beta}}
\newcommand{\bole}{\boldsymbol{\varepsilon}}

\newcommand{\boly}{\boldsymbol{y}}

\newcommand{\bolg}{\boldsymbol{\gamma}}
\newcommand{\bolth}{\boldsymbol{\theta}}

\newcommand{\bol}[1]{\boldsymbol{#1}}

\newcommand{\bone}{\mathbbm{1}}

\newcommand{\twonorm}[1]{\norm{#1}_2}
\newcommand{\onenorm}[1]{\norm{#1}_1}

\definecolor{dkgreen}{rgb}{0,0.6,0}
\definecolor{gray}{rgb}{0.5,0.5,0.5}
\definecolor{mauve}{rgb}{0.58,0,0.82}
\definecolor{BurntOrange}{rgb}{1,0.35,0}

\begin{document}

\title{Robust Sonar ATR Through Bayesian Pose Corrected Sparse Classification}
\author{John~McKay,~\IEEEmembership{Student Member,~IEEE,}
        Vishal~Monga,~\IEEEmembership{Senior Member,~IEEE,}
        and~Raghu~G.~Raj,~\IEEEmembership{Member,~IEEE}
\thanks{J. McKay and V. Monga are with the Department
of Electrical and Computer Engineering, Pennsylvania State University, University Park, PA
 e-mail:(John.McKay@psu.edu; VMonga@engr.psu.edu).}
\thanks{R. Raj is with the U.S. Naval Research Laboratory, Washington DC}
\thanks{A preliminary version of this work was presented at IGARSS 2016.}
\thanks{This work was supported by ONR Grant 0401531.}
}



\maketitle

\begin{abstract}
Sonar imaging has seen vast improvements over the last few decades due in part to advances in synthetic aperture Sonar (SAS).  Sophisticated classification techniques can now be used in Sonar automatic target recognition (ATR) to locate mines and other threatening objects.   Among the most promising of these methods is sparse reconstruction-based classification (SRC) which has shown an impressive resiliency to noise, blur, and occlusion.  We present a coherent strategy for expanding upon SRC for Sonar ATR that retains SRC's robustness while also being able to handle targets with diverse geometric arrangements, bothersome Rayleigh noise, and unavoidable background clutter.  Our method, pose corrected sparsity (PCS), incorporates a novel interpretation of a spike and slab probability distribution towards use as a Bayesian prior for class-specific discrimination in combination with a dictionary learning scheme for localized patch extractions.  Additionally, PCS offers the potential for anomaly detection in order to avoid false identifications of tested objects from outside the training set with no additional training required.
  Compelling results are shown using a database provided by the United States  Naval Surface Warfare Center.
\end{abstract}


\IEEEpeerreviewmaketitle

\section{Introduction}\label{sec:introduction}

Underwater mines are a significant concern for military and commercial endeavors.  When such an object is identified the task of removal is dangerous and costly, making the quality of the assessment extremely important.  Currently, automated underwater vehicles have emerged as powerful tools able to capture spacious underwater scenes with modern Sonar imaging techniques.  These machines are able to maneuver into places previously unreachable by manned vessels with no risk to operators.  Many are capable of implementing synthetic aperture Sonar (SAS), a technique that yields the highest resolution images in aqueous settings \cite{hayes2009synthetic,hansen2011introduction}.   

Given the sizable areas automated underwater vehicles can patrol as well as the quality of their images, the problem of automated mine identification has potentially never been as fertile as it is now.  While methods that work with direct human involvement have been common for Sonar ATR, the plethora of images makes a fully automated approach the only way to parse through each pixel in a timely fashion \cite{kriminger2015online,stack2011automation}.  This form of Sonar automatic target recognition (ATR), as it is known formally, came into existence three decades ago and has been improving steadily since.  Now, SAS has opened the field to sophisticated computer vision methods that were previously unfit for the low resolution underwater imagery.  This technique has allowed for geometric feature extraction \cite{isaacs2015sonar}, advanced shadow-based classifiers \cite{kumar2015robust}, and others.  SAS does unfortunately come with the potential for heavy noise, distinct blurring, and problematic target occlusion \cite{hayes2009synthetic,cook2009analysis}.  Therefore, modern techniques have had to be ready to handle substantial imperfections in order to perform reliably.

While Sonar ATR has been developing and improving, sparsity-based classification algorithms have carved a foot-hole into machine learning circles as one of the more robust options for target identification in the meantime.  The original, sparse reconstruction-based classification (SRC) \cite{wright2009robust}, has proven to be a state-of-the-art method in terms of facial recognition problems, even in settings of astronomical noise.  This relatively straightforward but powerful method and its variants have had success in identifying diseased tissue \cite{vu2015dfdl}, target recognition in Radar \cite{song2016sparse}, and handwriting analysis \cite{wagner2012toward}.  Many of the unavoidable issues in SAS image classification, including primarily noise, are ripe for localized sparsity-based approaches.

\textbf{Contributions:} The focus of this article is to effectively exploit sparsity as a prior in developing robust algorithms for classifying SONAR images. Our key contributions are:
\enum[1)]
\ii  \textbf{Bayesian Sparsity for Target Classification:} Recent work \cite{fandos2009sparse,kumar2012object,mckay2015discriminative,mckay2016localized} has shown promise for the use of SRC for SONAR ATR. We extend these ideas by using a novel spike and slab probability distribution construction as a Bayesian prior that can provide the discriminatory nuance necessary to discern targets in Sonar images. The use of such a sparsity inducing prior provides a more general sparse structure than the commonly used $\ell_0$ or $\ell_1$ measures and crucially, a way to capture class-specific information.

\ii \textbf{Locality to enable pose robustness and enhance discrimination:} A key shortcoming of SRC methods applied to Sonar is that they typically assume that training and test images (or suitably transformed versions) are consistent w.r.t the pose \cite{fandos2009sparse,kumar2012object,mckay2015discriminative} . We develop a local, patch-extraction method customized to handle pose variations in Sonar image capture that further incorporates a \emph{dictionary learning} scheme. We demonstrate that ATR at the level of local patches not only enables geometric robustness but local image features are also known to enhance recognition performance \cite{srinivas2014simultaneous}. We call our proposed technique as pose corrected sparsity (PCS).

\ii \textbf{Anomalous target identification:}  Many classification schemes are inherently forced to assign a label from their training set to any target even if it is a foreign object.  This poses a complication particularly in Sonar ATR settings where unknown seafloor targets may be worth further investigation.  To tackle this we propose a method to compare distributions of likelihood values from the training data to that of any test image to ensure any foreign object is deservingly flagged as an anomaly. 
 
\ii \textbf{Experimental Insights:} The proposed PCS is found to be meritorious over state of the art alternatives in two new scenarios: a.) when training Sonar imagery is limited, which is a very practical case because unlike many image classification problems in other domains \cite{wagner2012toward}, the number of example training images from each class are often constrained, and 2.) while most existing methods show reliable performance under additive Gaussian noise, we demonstrate benefits under the more realistic multiplicative Rayleigh noise which is more representative of distortions present in practical SONAR images \cite{hayes2009synthetic}.
\eenum

The rest of the paper is organized as follows:  Section \ref{sec:previousatr} presents the current state of Sonar ATR.  Section \ref{sec:pcs} provides the necessary background and technical details behind PCS including our novel Bayesian construction, the tailored refined sampling strategy, and the overall model design.  Section \ref{sec:anomaly} outlines our anomaly detection idea and how we suggest it to be implemented.  Section \ref{sec:experiments} is where we demonstrate compelling experimental results of PCS, including tests in limited training scenarios and noise robustness against other Sonar ATR methods.  Concluding remarks are presented in Section \ref{sec:discussion}.

\section{Previous Work in Sonar ATR}\label{sec:previousatr}
Mine hunting procedures can be broken down into two categories:  a ``search-classify-map''  (SCM) phase that picks through a scene to find potential targets and a ``reacquire-identify-neutralize'' (RIN) step that identifies targets and proceeds based on their perceived threat.  Work has generally  focusing on either just SCM or RIN with the former most interested in being fast and the latter in being accurate \cite{stack2011automation}.  Several innovative algorithms have been proposed for SCM, including those based on diffusion maps \cite{mishne2013multiscale}, Markov random field models \cite{reed2003automatic,mignotte1999three,goldman2005anomaly}, windowed area pixel intensity thresholds \cite{wang2015object}, and a locality-base graph models \cite{mishne2015graph}.  These schemes have the difficult task of locating mines even when faced with substantial background clutter, a common attribute of Sonar images.  Sand patterns, rocks, and debris can all confuse an object detection algorithm and they generally lead to an abundance of false alarms \cite{stack2011automation}.  Therefore, their value is in locating regions of interest (ROIs) where Sonar ATR algorithms can make assessments.  Note that in the following we will refer to Sonar ATR in its capacity to identify that which lies in a ROI.  In a broader sense, Sonar ATR encompasses both SCI and RIN, but we are narrowing it down for clarity.

Sonar ATR research has developed several options for classifying targets in ROIs.  Template matching and feature description schemes have been the most commonly pursued.  The former involves sophisticated simulations of targets under multiple scenarios to generate image patterns similar to those found in practice, including the type of shadows that can be found from differently shaped objects.  From here, correlation statistics or feature extractions are used on targets to determine which of the templates is the best match and a classification is made \cite{myers2010template} \cite{groen2010model}.  Template matching methods require software that can effectively replicate the physics behind Sonar imaging, limiting their general usability.

Feature description algorithms have become increasingly popular.  Here, Sonar images go through feature transformations in bulk, a selection phase to isolate the most discriminative descriptors, and then they get tested against a trained model.  While there is no one set of features that has proven optimal for Sonar ATR, recent algorithms have had success with statistical features based on the shapes of targets, their shadows, and environmental effects \cite{fandos2011optimal,isaacs2011laplace,kumar2015robust,isaacs2015sonar,williams2014exploiting}.  Other algorithms have been designed to pick features in a manner most conducive for Sonar ATR as well \cite{fei2015contributions}.  After extraction and selection, classification based on the resulting features has taken several routes including neural nets \cite{dobeck1997automated} and adaboost models \cite{isaacs2011signal} \cite{isaacs2015sonar}.  Fusion-based approaches that combine classifiers have been of much recent interest \cite{azimi2000underwater}  \cite{kumar2015robust}.

A key new direction has been applying sparse classification techniques for Sonar ATR.  Kumar \emph{et al} uses a classification technique similar to SRC that we cover in detail in Section \ref{subsec:src}, except with an elastic-net solution to a feature selection problem \cite{kumar2012object}.  The authors extract Zernike moment statistics from ROIs and use these in their sparsity problem.  Fandos \emph{et al} also use a straight-forward SRC framework to classify well-aligned targets in ROIs, though they did not try to correct for noise, blur, or ill-posed targets  \cite{fandos2009sparse}.

In the next section, we describe a novel Sonar ATR algorithm that does not require any feature transformations and extends the success of sparsity-based classifiers for classifying sonar images.
Two significant emphasis points of our work, not pursued before include:  1.) the use of Bayesian sparse priors that enable parameter design to capture class specific behavior, and 2.) a locality based approach involving patch extraction and refinement, which enables geometric robustness.

\section{Proposed Work: Bayesian Pose Corrected Sparsity for Sonar ATR}\label{sec:pcs}
The challenges of Sonar ATR, none the least of which being reliability, make sparse reconstruction-based methods a natural application.  We present a sparsity based classification approach, Pose corrected Sparsity (PCS) that is specifically tailored for Sonar ATR. PCS utilizes sparsity-enforcing Bayesian priors and intensity-focused, intelligently filtered patch extraction coupled with discriminatory dictionary learning to robustly classify several classes underwater objects.

\subsection{Sparse-Reconstruction Classification}\label{subsec:src}
Sparsity refers to how well populated a vector is with non-zero entries.  A sparse vector is one that contains mostly zeros. Generally, the power of sparsity comes in the way of an optimization constraint that encourages vectors to conservatively allot non-zero entries.  Compressive sensing was one of the first image processing fields to investigate sparsity in detail.  Sparsity constraints have produced state-of-the-art image compression using far fewer samples that Nyquist-Shannon suggested and incredible progress in image denoising, to name just two successful applications  \cite{candes2008introduction}.  

Sparsity in compressive sensing is implemented within straight-forward linear models.  Given a matrix $X\in\reals^{N\times M}$, the \emph{dictionary}, composed of wavelet, Fourier, or other basis elements and $\boly\in\reals^N$ representing the vectorized test image, the coefficients $\bolb^\star\in\reals^M$ are modeled as
\beqa\label{eq:cs}
\boly = X\bolb^\star + \bole
\eeqa
where $\bole\sim\mathcal{N}(\bol{0},\sigma^2I)$ represents white Gaussian noise.  This construction, similarly to a regression model in statistics, is commonly solved with an optimization problem involving a penalization term
\beqa\label{eq:min1}
\bolb^\star =\arg\min\limits_{\bolb\in\reals^M} \twonorm{\boly - X\bolb}^2 +\lambda \text{\bf pen}(\bolb)
\eeqa
The notion of sparsity is typically enforced by letting the penalty function be the zero pseudo-norm, $\norm{\cdot}_0$ - that is - make it the count of non-zero entries of its vector input (the ``pseudo'' comes from the fact that it does not satisfy the triangle inequality and absolute homogeneity).  In this form, the problem becomes
\beqa\label{eq:zpn}
\bolb^\star =\arg\min\limits_{\bolb\in\reals^M}\twonorm{\boly-X\bolb}^2 +\lambda\norm{\bolb}_0
\eeqa
Why is sparsity so useful?  The penalty function forces the optimization problem to find the most influential basis elements in the columns of $X$ to reconstruct $\boly$.  This frugal take on coefficient values leads to solutions that involve only but a few entries of $\bolb$ that also in turn reflect the contribution of their corresponding basis elements, an essential factor as we will see for classification.

The work of \cite{wright2009robust} is the first to adapt sparsity constraints towards image classification with their sparse reconstruction-based classification algorithm (SRC).  The authors take the same framework as \eqref{eq:zpn} for classifying an image $\boly$ except they define $X$ not as a collection of wavelets or Fourier basis elements but as a concatenation of training images from several classes.  That is, consider a set of training images $\{\bolx_{i,k}: i=1,\dots,n\,\,\,k=1,\dots,K\}$ where $K$ denotes the number of classes.  The dictionary $X\in\reals^{N\times M}$ for $M=nK$ is defined as
\beqa\label{eq:dic}
X=\begin{pmatrix} 
X_1 & \cdots & X_K\\
\end{pmatrix}
\eeqa
where each of $X_k\in\reals^{N\times n}$ is a concatenation of the training images corresponding to each of the $K$ classes
\beqa\label{eq:xk}
X_k=\begin{pmatrix} \smallb & & \smallb\\
\bolx_{1,k} & \cdots & \bolx_{n,k}\\
\smallb&&\smallb
\end{pmatrix}
\eeqa
In finding the $\bolb$ that best reconstructs $\boly$, the coefficients can be parsed based on their associated classes and then used to figure out which class alone best reconstructs the test image. In other words, the class $k^\star$ given to $\boly$ is found via
\beqa\label{eq:class}
k^\star = \arg\min_{k\in 1,\dots K} \twonorm{\boly-X\bol{\delta}_k(\bolb^\star)}
\eeqa
where $\bol{\delta}_k$ is a vector-valued function that keeps the entries of $\bolb^\star$ corresponding to class $k$ and ``zeros-out'' the others.

The sparse optimization problem of \eqref{eq:zpn} is \emph{fundamental} to the success of SRC.  Wright \emph{et al} \cite{wright2009robust} show in their work that a significant merit of SRC over the established body of traditional image classification methods is that SRC exhibits unprecedented levels of robustness to noise and occlusion, i.e. even as test images are noisy or occluded the sparse code as solved using (\ref{eq:zpn}) remains largely invariant.

\noindent \textbf{Customizable sparse structure using Bayesian Spike and Slab Priors:}  Note that sparsity is enforced using the term $\norm{\cdot}_0$.  This pseudo-norm makes \eqref{eq:zpn} a non-convex NP-hard problem, so Wright \emph{et al} offered an $\ell_1$ relaxation approach \cite{wright2009robust}.  This is a limiting tactic to handling the $\ell_0$ pseudo-norm and removes a great deal of potential nuance, especially in terms of class-specific tailoring.  We develop a probabilistic end-around to this issues:  a \textbf{spike and slab prior} for $\bolb$.  To illustrate this, we first start with the maximal likelihood understanding of SRC problem:
\beqa\label{eq:bayes}
\bolb=\arg\max\limits_{\bolb} P(\bolb\,|\,\boly)=\arg\max\limits_{\bolb} P(\boly\,|\,\bolb)P(\bolb)
\eeqa
Building off of our assumptions behind \eqref{eq:cs}, we see that
\beqna
\boly\,|\,\bolb\sim\mathcal{N}(X\bolb,\sigma^2 I)\propto \exp\lp -\frac{1}{2\sigma^2}\twonorm{\boly-X\bolb}^2\rp
\eeqna
meaning an equivalent problem statement for \eqref{eq:bayes} is
\beqa\label{eq:lnbayes}
\bolb^\star &= \arg\max\limits_{\bolb}-\frac{1}{2\sigma^2}\twonorm{\boly-X\bolb}^2-\ell(\bolb)\\
&= \arg\min\limits_{\bolb}\frac{1}{2\sigma^2}\twonorm{\boly-X\bolb}^2+\ell(\bolb)\\
& =\arg\min\limits_{\bolb}\twonorm{\boly-X\bolb}^2+2\sigma^2\ell(\bolb)
\eeqa
where $\ell(\bolb)$ is the log prior probability distribution given to $\bolb$.

Notice that the $\ell_1$ relaxation to \eqref{eq:zpn} really imparts a Laplacian distribution upon $\bolb$ under this framework as
\beqa\label{eq:Laplacian}
\bolb\sim \tau\exp(-2\tau\onenorm{\bolb})
\eeqa
yields a one norm penalty with $\lambda=2\sigma^2\tau$ when applied to \eqref{eq:lnbayes}.  Unfortunately, the Laplacian distribution does not explicitly incur sparsity and lacks any flexibility to incorporate additional information about the $\bolb$ into the problem.  It is too rigid for any further discriminatory action.

A more pliable prior that does encourage sparsity directly is the spike and slab distribution.  This mixture distribution models each entry of $\bolb$ individually with
\beqa\label{eq:bolb}
\beta_i \sim (1-\gamma_j)\bone_0 + \gamma_j\mathcal{N}(0,\kappa^2)
\eeqa
where $\gamma_j$ is binary, $\bone_0$ is an indicator function such that
\beqna
\bone(x)=\begin{cases}
1,&x=0\\
0,&x\neq 0
\end{cases}
\eeqna
and $\kappa^2$ is the variance for non-zero elements of $\bolb$.  This flips between two outcomes depending on the value of $\gamma_j$:  a guaranteed value of zero if $\gamma_j$ is one (the spike) and Normally distributed values if $\gamma_j$ is zero (the slab).  This ``on-off switch'' parameter $\bol{\gamma}$ is what directly models sparsity and by taking each of its entries as independent Bernoulli trials, we can model $\bolg=[\gamma_1\cdots\gamma_M]^T$ as
\beqa\label{eq:gamma}
\bolg\,|\,\bolth \sim \prod\limits_{j=1}^M \theta_j^{\gamma_j}(1-\theta_j)^{1-\gamma_j}
\eeqa
where $\bolth=[\theta_1\cdots\theta_M]^T\in[0,1]^M$ is the probability that any entry is non-zero.

Putting all the pieces together, the spike and slab prior for $\bolb$ is \cite{ishwaran2005spike}:
\beqa\label{eq:spikeslab}
\boly\,|\,\bolb,\sigma^2 &\sim \mathcal{N}(X\bolb,\sigma^2I)\\
\bolb\,|\,\bolg,\kappa^2&\sim\prod\limits_{j=1}^M(1-\gamma_j)\bone_0 + \gamma_j\mathcal{N}(0,\kappa^2)\\
\bolg\,|\,\bolth &\sim\prod\limits_{j=1}^M \theta_j^{\gamma_j}(1-\theta_j)^{1-\gamma_j}
\eeqa
If we make the appropriate log transformations and simplifications, the above framework results in the following maximal likelihood construction \cite{mousavi2015iterative}:
\beqa\label{eq:lnbayesss}
\arg\min\limits_{\bolb,\bolg}\twonorm{\boly-X\bolb}^2 +\alpha \twonorm{\bolb}^2 + \sum\limits_{i=1}^M\xi_i \gamma_i
\eeqa
with $\alpha=\tfrac{\sigma^2}{\kappa^2}$ and $\xi_i=\ln(\tfrac{2\pi\kappa^2(1-\theta_i)^2}{\theta_i^2})\sigma^2$.  Since each entry $\gamma_i$ is binary, reflecting whether or not the corresponding $\beta_i$ value is non-zero, a solution to \eqref{eq:lnbayesss} is such that the reconstruction error between $X\bolb$ and $\boly$ is minimal with respect to the most conservative choices of non-zero values of $\bolb$ along with their different penalties.  What if those penalties were equal, i.e. $\xi_i=\xi$ for all $i$?  In this case, notice that by construction, $\norm{\bolg}_0=\norm{\bolb}_0$, so we can say
\beqa\label{eq:lnbayessst}
&\arg\min\limits_{\bolb,\bolg}\twonorm{\boly-X\bolb}^2 +\alpha \twonorm{\bolb}^2 + \xi \sum\limits_{i=1}^M\gamma_i\\
& = \arg\min\limits_{\bolb,\bolg}\twonorm{\boly-X\bolb}^2 +\alpha \twonorm{\bolb}^2 + \xi \norm{\bolg}_0\\
& = \arg\min\limits_{\bolb}\twonorm{\boly-X\bolb}^2 +\alpha \twonorm{\bolb}^2 + \xi \norm{\bolb}_0
\eeqa
The case of equal $\xi_i$ terms reveals a cost function similar to that of \eqref{eq:zpn} except with additional controls on the non-zero entries of $\bolb$.  When do we have such a case of a uniform penalty?  When the probability $\theta_j$ is equal amongst all entries.  Thus, the spike and slab prior gives us a crucial insight of our sparse reconstruction problem:  the typical construction \eqref{eq:zpn} imparts a non-discriminative sense of sparsity towards every coefficient value.

What if $\bolth$ did not have equal entries?  That is, what if we allowed for different sparsity parameters and looked to solve a problem more akin to \eqref{eq:lnbayesss}?  Doing as such is definitely plausible, but daunting.  Figuring out appropriate calculations for every entry involves considerable cross-validation, not to mention that traditional MCMC approaches to learning parameters would reduce the reconstruction error and not necessarily search for values apt for classification.  To alleviate this, we suggest \textbf{class-specific sparsity values}.  Given $K$ classes, we assume each to have their own inherent sparseness so that we solve
\beqa\label{eq:lnbayesk}
\arg\min\limits_{\bolb,\bolg}\twonorm{\boly-X\bolb}^2 +\alpha \twonorm{\bolb}^2 + \sum\limits_{k=1}^K\xi_k\sum\limits_{i\in\mathcal{I}_k}\gamma_i
\eeqa
where $\mathcal{I}_k$ is the index set of $\bolb$ corresponding to class $k$.  This novel construction offers flexibility to impact classifications made using reconstruction error such as with \eqref{eq:class}.  In a sense, \eqref{eq:lnbayesk} allows us to dampen the effect of any class with respect to the others by increasing its relative sparseness, meaning that fewer elements of the dictionary associated with this class can contributed to an overall reconstruction.  By doing so, if there is a class that contains several varied characteristics that make a classification confusing, this dampening effect can make it so that only the tests that closely match to the most salient discriminatory features of that class are ultimately sufficient.

\begin{figure}[t]
\centering
\includegraphics[width=.93\columnwidth]{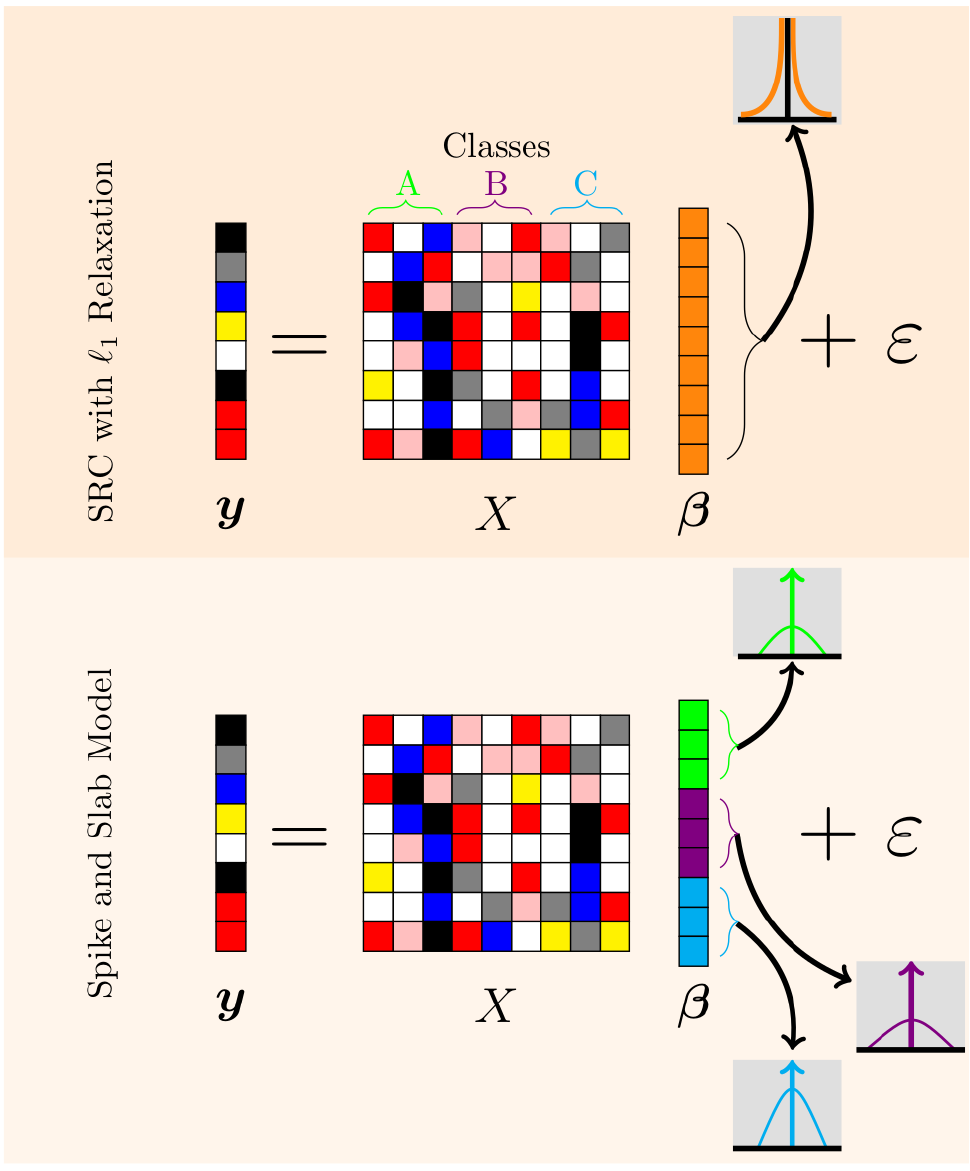}
\caption{Basic idea behind SRC with $\ell_1$ relaxation vs. the spike and slab model.  The maximal likelihood interpretation of \ref{eq:lnbayes} shows that $\ell_1$ results in an exponential distribution underlying $\beta$.  Our spike and slab model provides a flexibility so that each class gets its own distribution settings.}\label{fig:spikeandslab}
\end{figure}

 To solve \eqref{eq:lnbayesk}, the most adaptable method is the iterative convex refinement (ICR) algorithm presented in \cite{mousavi2015iterative}.  Typically, we focus on choosing different $\alpha$ and $\xi_k$ values instead of the raw $\theta_k$, $\kappa$, and $\sigma^2$.  In doing so, we end up ultimately with the spike and slab model shown in \ref{fig:spikeandslab} where instead of the Laplacian distribution underpinning $\bolb$, we each class with their own ``on-off switch'' Gaussian distribution, implicit by the choice of $\alpha$ and $\xi_i$.

\subsection{Locality Through Patch Extraction \& Dictionary Learning}\label{subsec:patchesanddl}

While SRC and our spike and slab offer the potential for robust classification, they are inherently unable to tackle problems whose test images are not geometrically aligned with the training.  Indeed, if an object is detected within some windowed region but lies off-center with respect to the training, it is quite possible that a misclassification will occur regardless of how similar the actual object is to the training.  In the following section, we will detail measures that do not just alleviate this issue but also equip our spike and slab approach with heightened locality to further add a discriminatory element.

The problem of sparsely constrained classification and pose has been investigated with much of the focus coming from facial recognition settings.  Several algorithms have been developed that depend on extensive control of the training images to create multiple different capture points as to construct image sets \cite{cevikalp2010face} \cite{kokiopoulou2010graph} or shape manifolds for better distance estimations \cite{fukui2005face}.  These approaches are infeasible for Sonar settings where directing an AUV to capture an allotment of different looks of a single target is expensive and difficult in its own, let alone for every type of mine-like object.  Additional work has been done into pose-normalization where models are trained on frontal views and tests go through a rotational transformation to fit the arrangement of their training \cite{zhang2012pose}.  This strategy is similar in some ways to the template matching schemes described in section \ref{sec:previousatr} but instead of trying to create a flexible training set, the pose-normalization attempts to correct the testing image.  Unfortunately, this too fails to translate easily for Sonar given its training requirements as well as the difficulty of rotating a target, especially amongst rugged backgrounds.

Another approach that has proven useful comes from a patch-extraction technique \cite{zhang2015survey}.  This not only circumvents alignment issues but also allows for differing target window sizes.  It works as follows:  given the set of training images $\bolx_{i,k}$ from before, we take $P$ patches from each yielding the expanded collection
\beqa\label{eq:patches}
\{\bolx_{i,k}^p\in\reals^b\,:\, i=1,\dots n\,\,\,k=1,\dots, K\,\,\,p=1,\dots,P\}
\eeqa 
for $b=b_1b_2$.  The patches assume the label given to their original training image.  For the dictionary $X\in\reals^{b\times MP}$ we again concatenate our training elements so that
\beqa\label{eq:dpatch}
X    =\begin{pmatrix}
\dual X_1 & \cdots &\dual X_K\\
\end{pmatrix}
\eeqa
where $\dual X_k\in\reals^{b\times nP}$ is
\beqa\label{eq:dpatch}
\dual X_k   =\begin{pmatrix}
 \smallb & & \smallb & & \smallb\\
 \bolx_{1,k}^1 & \cdots &\bolx_{n,k}^1&\cdots &\bolx_{n,k}^P\\
  \smallb & & \smallb & & \smallb
\end{pmatrix}
\eeqa
Now if we were to use every patch, the size of the dictionary would be enormous and potentially untenable for any reasonable machine.  Thus, a still sizable random sample of patches is used which has been shown not to impact overall resulting classification rates noticeably \cite{vu2015dfdl}.  As well, note that this patch extraction scheme explicitly accounts for translation invariance.  In Sonar where the angle at which an object is captured can dramatically change the resulting image, such an invariance is preferable \cite{mckay2015discriminative}.  \cite{srinivas2011sparsity} shows that a patch-based scheme can have a slight rotational robustness, which is appropriate for Sonar.  For the following, when we remark about ``pose correction" we are referring to an algorithm that has invariance in this manner useful for Sonar.

To classify an image $\boly$, we break it into $C$ patches $\boly_1,\dots,\boly_C\in\reals^b$ and find model coefficients $\bolb_1,\dots,\bolb_C\in\reals^{MP}$ accordingly using the spike and slab prior construction of the $\ell_0$ problem.  From here we use a strategy from ensemble classification to make an assessment of $\boly$ \cite{kittler1998combining}.  For each of the coefficients and test patches, the associated residual metric we use is defined as
\beqa\label{eq:residuals}
r_{i,k} =\frac{1}{\twonorm{\boly_i-X\bol{\delta}_k(\bolb)}}
\eeqa
Classification $k^\star$ is made with a simple maximization of log probabilities based on these terms, i.e.
\beqa\label{eq:choosek}
k^\star = \arg\max\limits_k \ln \lp \prod\limits_{i=1}^{C}\frac{r_{i,k}}{R_i} \rp \text{ where } R_i = \sum\limits_{i=1}^K r_{i,k}
\eeqa

\begin{figure}[t]\centering 
\includegraphics[width=.324\textwidth]{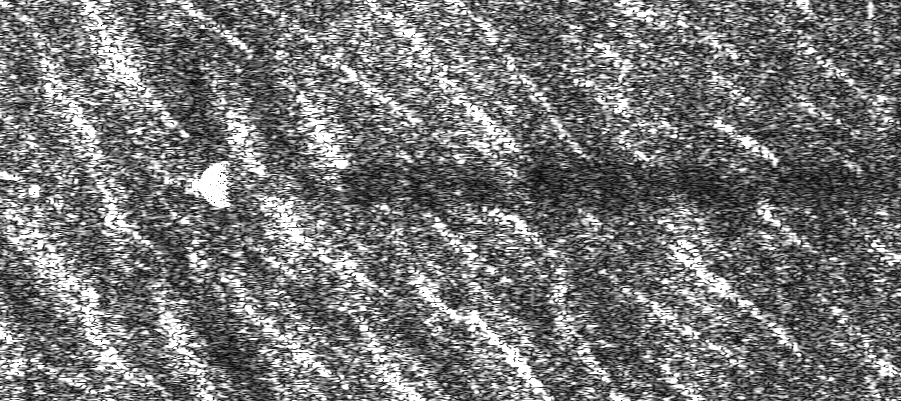}
\includegraphics[width=.145\textwidth,trim={1.6cm 1.6cm 1.6cm 1.6cm},clip]{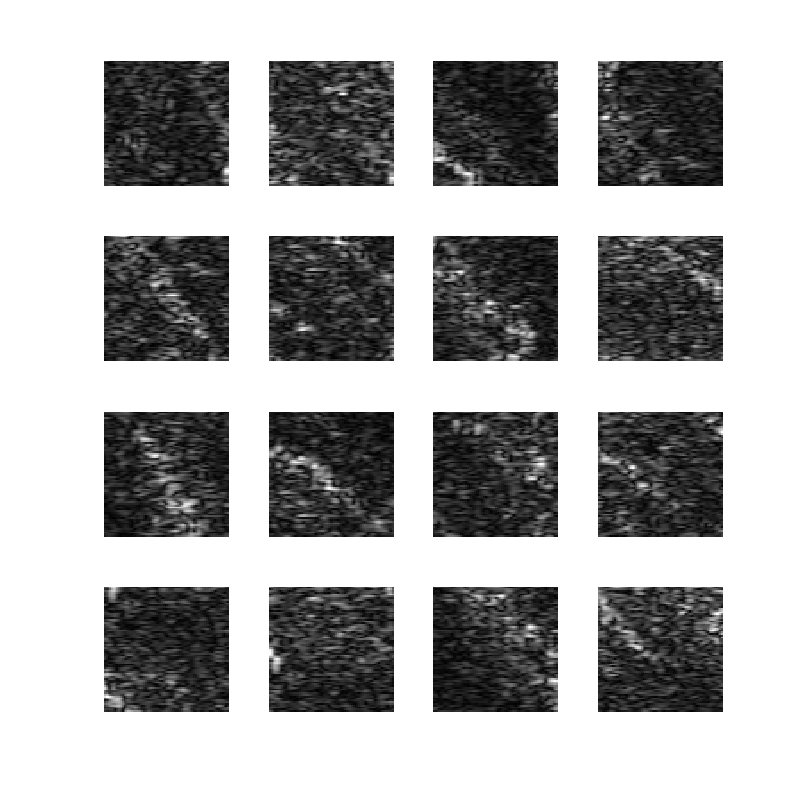}\\\vspace{-.3cm}\rule{8.5cm}{1pt}
\includegraphics[width=.324\textwidth,trim={2.25cm 1.1cm 2.25cm .6cm},clip]{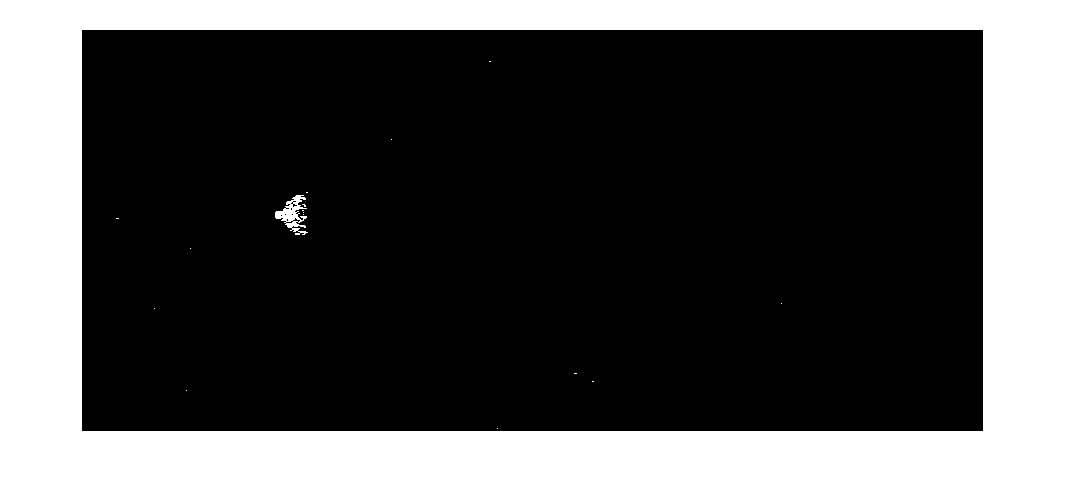}
\includegraphics[width=.145\textwidth,trim={1.6cm 1.6cm 1.6cm 1.6cm},clip]{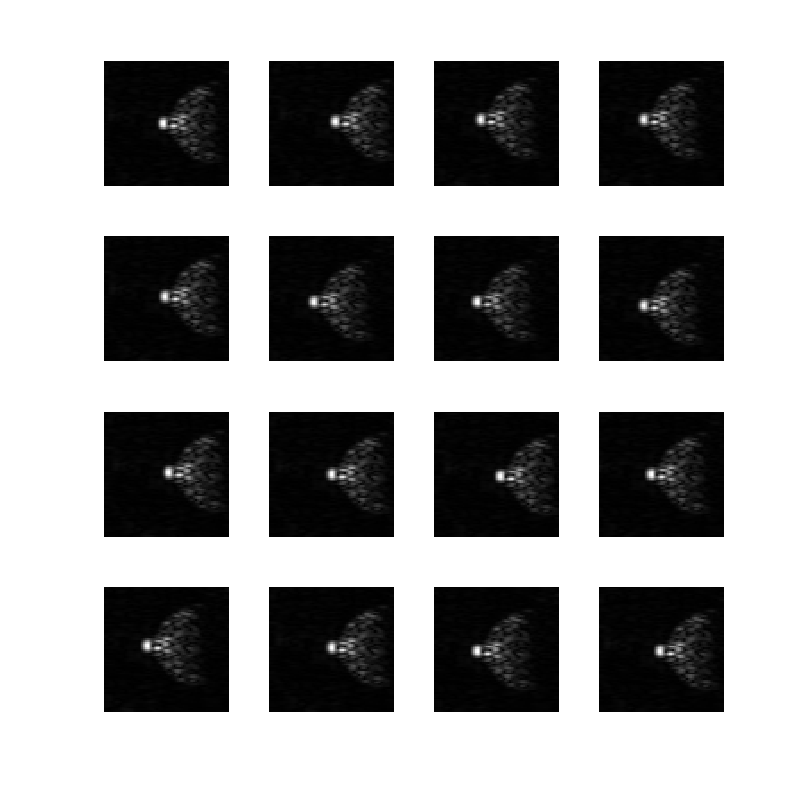}
\caption{SAS image of a sphere with random patches (top) and filtered image (white represents the surviving magnitude values) with random patches containing these regions (bottom).}\label{fig:intensity}
\end{figure}

At this point, we have a functional but still ill-prepared method for Sonar ATR.  Letting patches be collected from anywhere within a windowed region may let background elements seep into the dictionary which can mislead the classifier and impact quality.  Further, there is no assurance that the patches that do include the mine are necessarily helpful; local neighborhoods of different targets may been exceedingly similar.  Even though our spike and slab strategy help discriminate between classes that resemble each other, a filtration of patches that advances differences can only help.  For these reasons, we present two ways to handle these issues:  intelligent stratified sampling and dictionary learning.

\begin{figure}[t]\centering 
\begin{subfigure}[t]{1\columnwidth}\centering
\includegraphics[width=.32\columnwidth,trim={4cm 5cm 0 0},clip]{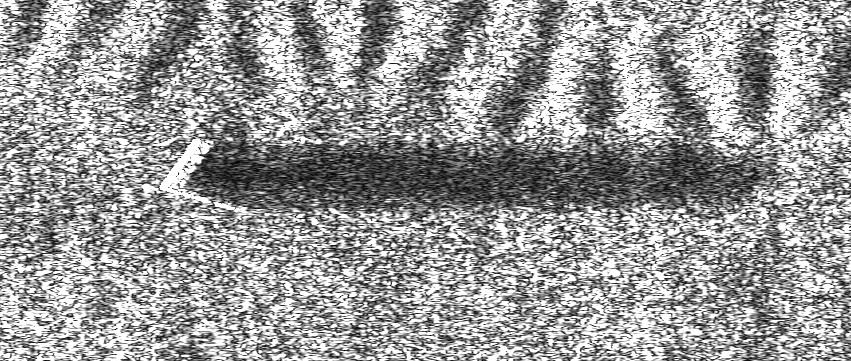}
\includegraphics[width=.32\columnwidth,trim={4cm 5cm 0 0},clip]{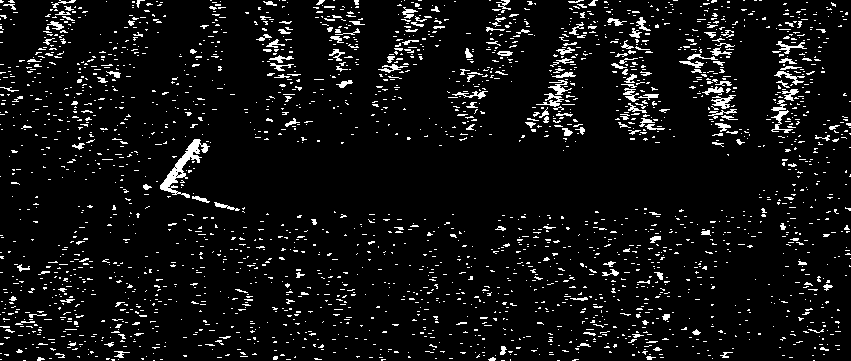}
\includegraphics[width=.32\columnwidth,trim={4cm 5cm 0 0},clip]{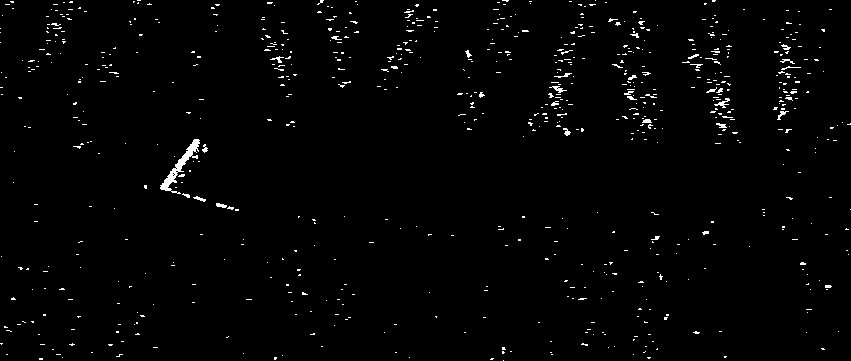}
\caption{Example SAS image with intensity thresholding for stratified patch extraction regions.  Left is the original image of a block and going right represents higher thresholded regions.}\label{fig:intensity2}
\end{subfigure}
\begin{subfigure}[t]{\columnwidth}\centering
\includegraphics[width=.23\columnwidth]{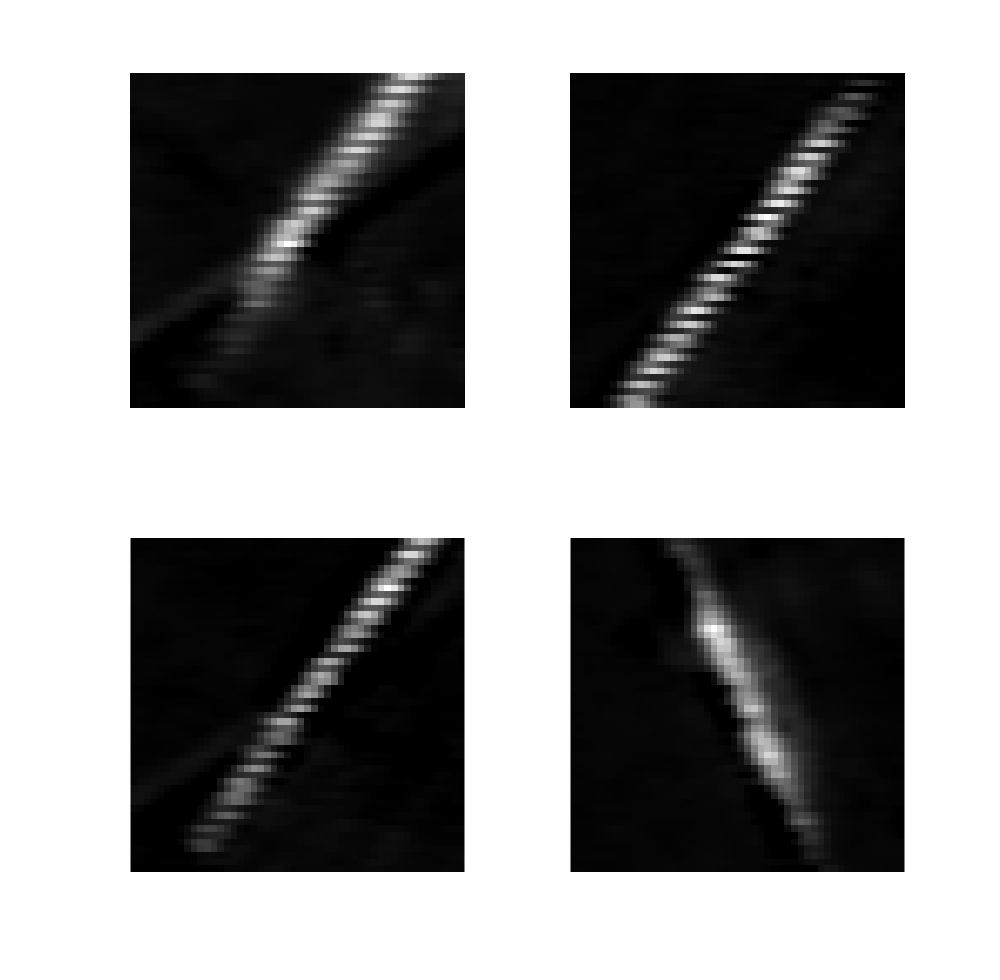}
\includegraphics[width=.23\columnwidth]{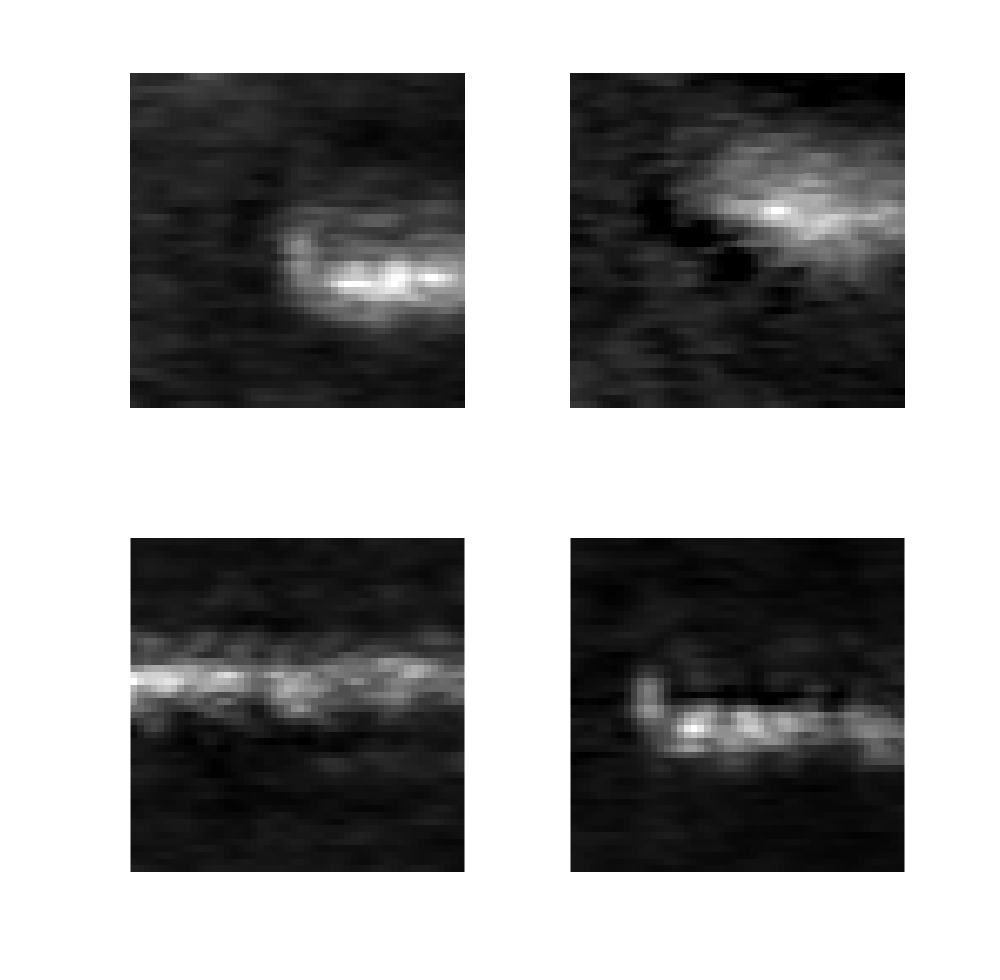}
\includegraphics[width=.23\columnwidth]{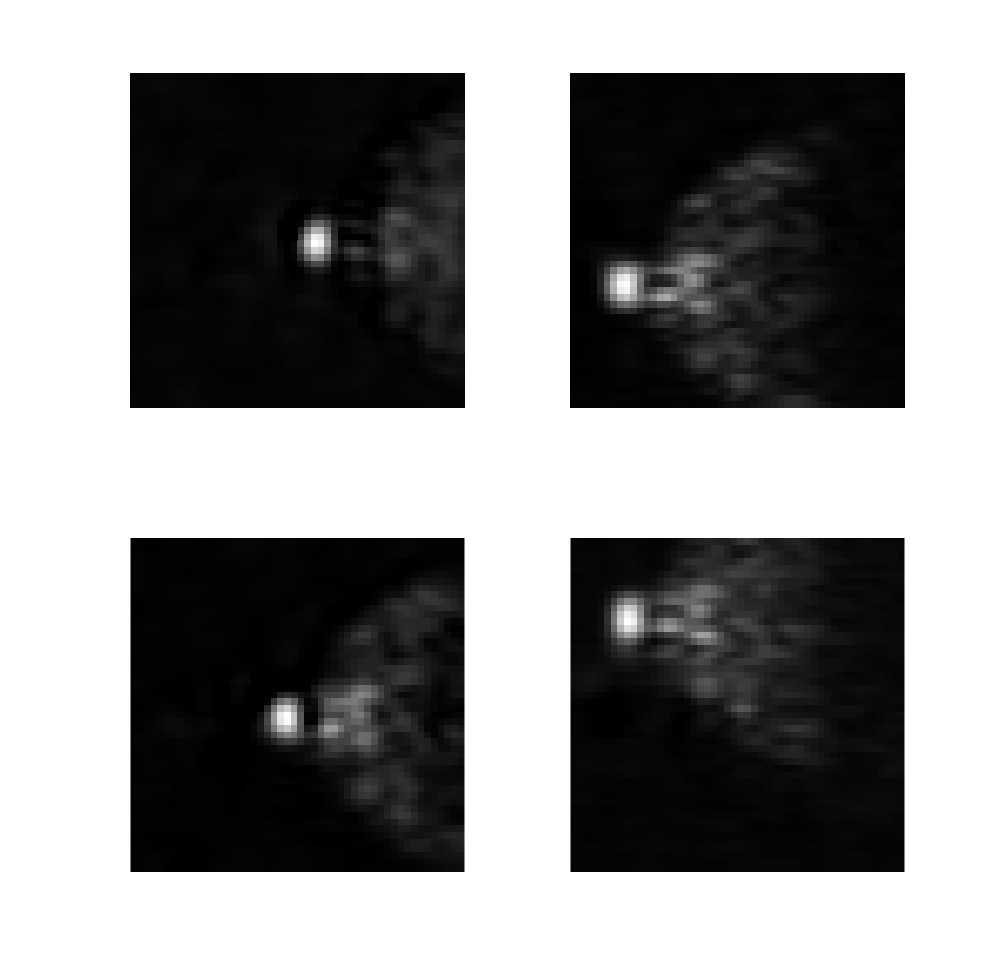}
\includegraphics[width=.23\columnwidth]{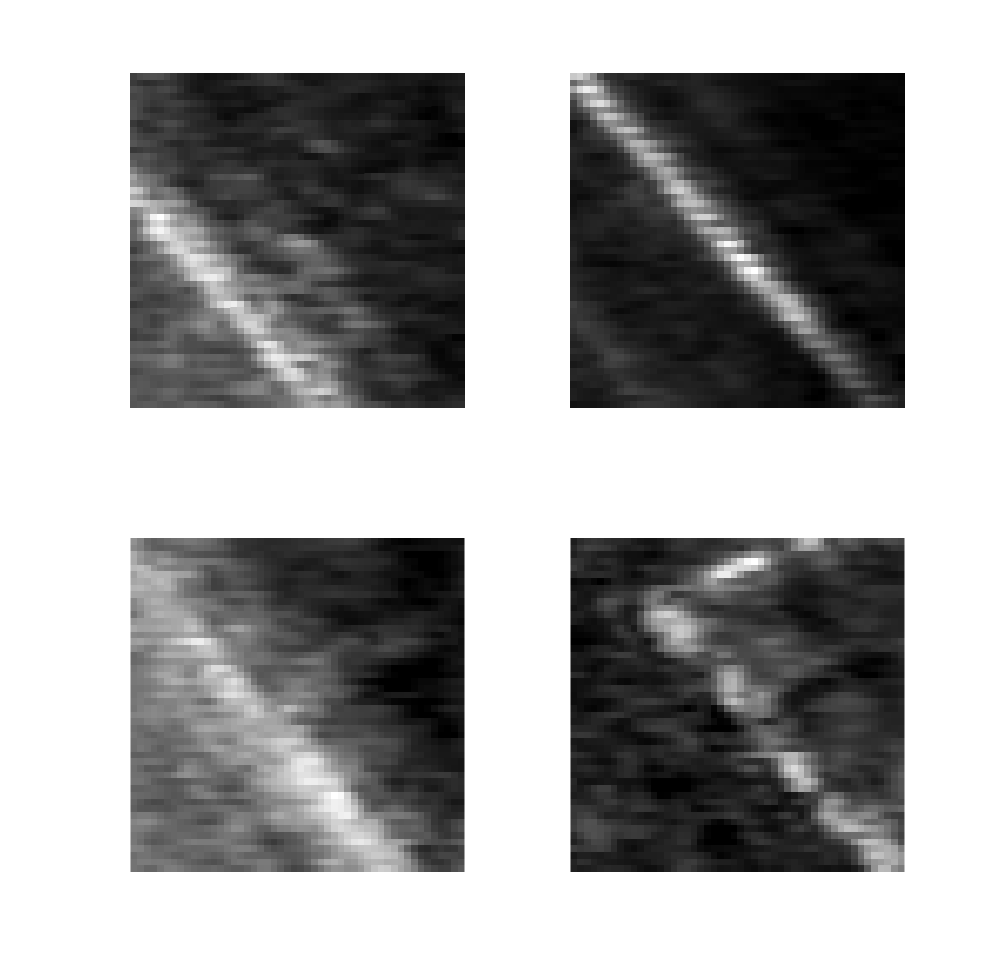}
\caption{Example learned patches by DFDL algorithm used in PCS.}\label{fig:learns}
\end{subfigure}
\caption{Problems with thresholded SAS images and example learned patches showing little issue.}
\end{figure}

There is a simple but powerful phenomenon in Sonar:  targets almost always contain regions of high intensity pixel values \cite{wang2015object}.  We can parlay this aspect into a smart sampling technique for patches by ensuring that each one contains pixel values above a certain threshold.  This does not preclude certain background clutter from still contaminating the collection of patches, but it does greatly improve the chances of useful information being gleaned from an extraction.  Additionally, this is a relatively quick and easy stratification to make.  Figure \ref{fig:intensity} demonstrates the effectiveness of this simple fix.  The original SAS image contains a spherical object that we ultimately are interested in classifying.  The patches taken without guidance are entirely just that of background given that the overwhelming area is mostly just that and the object takes up relatively little space.  Then, with just a modest thresholding that requests the bottom twenty percent of pixels be disregarded, we see that the target is almost the only region that survives and the selected patches reflect this!  Again, this easily implemented stratified sampling works incredibly well in withering down the pool of available patches - but it is not ultimately "perfect."  The region depicted in Figure \ref{fig:intensity2} shows how prominent background like the wavy sea-floor depicted can make it through to the filtered regions.  Thus, while this can dramatically reduce the number of useless patches, it alone is not enough to ensure a proper collection.  For this reason we implement a dictionary learning scheme.

To get a really discerning model, we need to refine our set of patches further so that the resulting dictionary can be augmented by the spike and slab prior - instead of completely reliant upon it.  Dictionary learning algorithms can do just that.  Dictionary learning refers to any framework designed to filter down the basis elements making up a dictionary to only the most essential, minimal set.  \cite{aharon2006ksvd} demonstrated the first method for compressive sensing and, since then, many algorithms have come forth for classification purposes in order to improve performance and limit computational effort.

Recalling that we are concerned with distinguishing between potentially similar classes, one way to consider learning a dictionary is to isolate the elements that both inspire inter-class heterogeneity and encourage intra-class homogeneity.  In this context, discriminative feature dictionary learning (DFDL) does just that for sparsity-based classification.  Formally, if we let $D$ be our learned dictionary, $\classin(\cdot)$ be a matrix of just the columns associated with class $k$ for the input, and $\classout(\cdot)$ be a matrix of all but the columns of class $k$ for the input, then DFDL works by solving
\beqa\label{eq:dfdl}
D = &\arg\min\limits_{X}\bigg\{ \frac{1}{n}\min_{\scriptscriptstyle\norm{\classin(B)}_0\leq L}\norm{\classin(Y)-X\classin(B)}_F^2 \\
&-\frac{\rho}{\bar n_k}
\min\limits_{\scriptscriptstyle\norm{\classout(B)}_0\leq L}\norm{\classout(Y)-X\classout(B)}_F^2\bigg\}
\eeqa
where $Y$ a matrix whose columns are the training samples, $B$ is the matrix of SRC coefficients corresponding to each of the columns of $Y$, $\norm{\cdot}_0$ represents the matrix zero pseudo-norm which counts the number of non-zero entries in each row, and $L$ is a learned sparsity level.  DFDL, with its explicit class-discriminatory construction, is a powerful tool that can be used within a sparsity based classification framework. A recent demonstration of the application of SRC using DFDL to the problem of identifying diseased areas in large images of different bodily tissues has found notable success by outperforming other dictionary learning schemes \cite{vu2015dfdl}.  Given these compelling results and our own success, we found this to be a suitable dictionary learning approach for our task.  To see example patches learned from DFDL, consult Figure \ref{fig:learns}.

\begin{table}[b]\centering
    \begin{tabular}{|c|c|c|c|c|}\hline
        \textbf{Training} & \textbf{1 Patch Classification} & \textbf{PCS}\\\hline
380.27s & 3.51s & 62.54s  \\\hline
     \end{tabular}
    \caption{Times for an example PCS problem.  $50\times 50$ patches were used for training and testing with 40 images per class being the basis for the PCS model.  The PCS implementation involved 17 patches from a test image.}\label{tab:cost}
\end{table}

\subsection{Sonar ATR with Pose Corrected Sparsity}\label{subsec:pcs}

\begin{figure}[t]\centering
\includegraphics[width=.94\columnwidth]{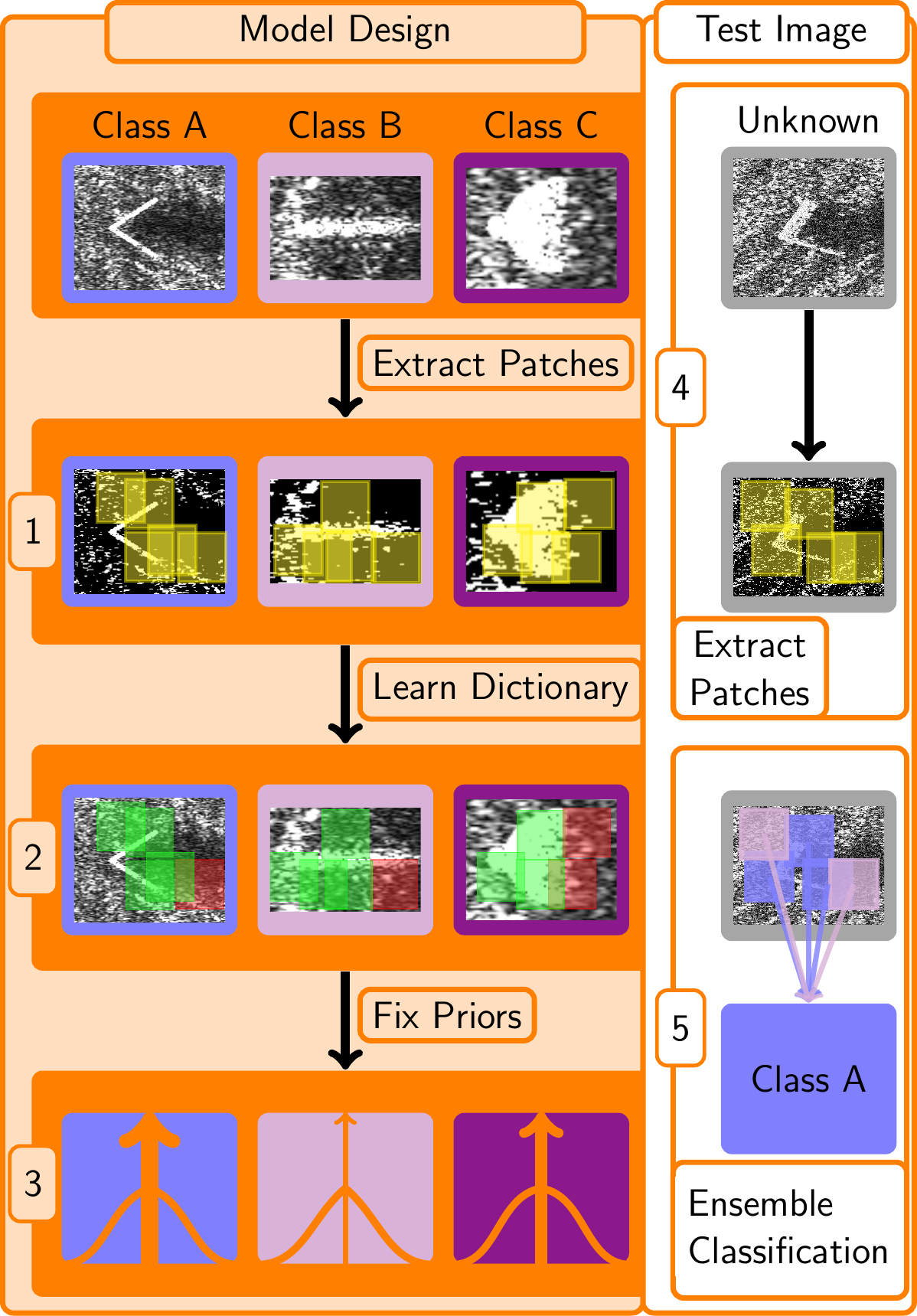}
\caption{Diagram of PCS for Sonar ATR.  Model construction goes as follows:  (1) sets of training images patches are sampled from high intensity regions which are then (2) refined according to the DFDL algorithm concentrate and condense our dictionary, which is used with re-sampled training patches for (3) cross-validation to estimate appropriate class-specific parameters for our spike and slab prior.  To test an image, (4) we extract patches from high intensity regions and (5) classify each one according to the learned dictionary, making a classification based on an aggregation of the reconstruction errors.}\label{fig:pcs}
\end{figure}

Pose corrected sparsity (PCS) is our name to the algorithm that incorporates the spike and slab Bayesian concept with the patch-extraction, dictionary learning set up to yield a robust, flexible classification algorithm.  The diagram in Figure \ref{fig:pcs} depicts the specific flow of PCS for an example three class problem.  

\begin{figure*}[t]\centering
\includegraphics[width=.99\textwidth]{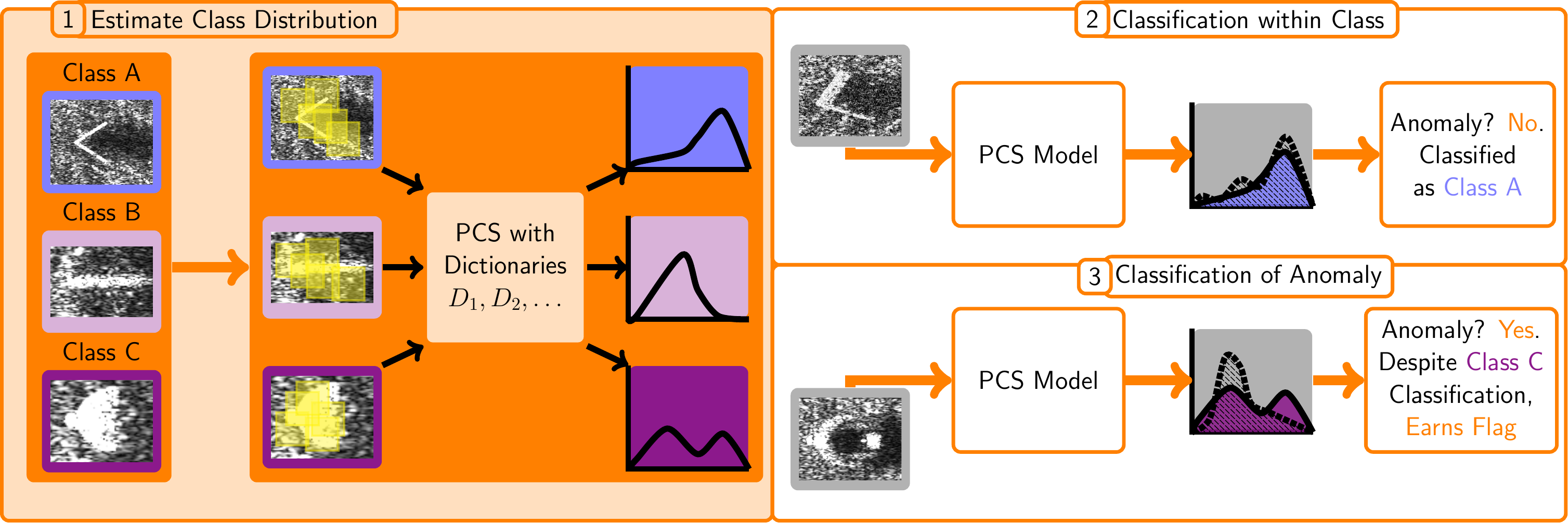}
\caption{Anomaly detection derived from PCS.  This depicts: (1) developing the class-specific residual distributions during cross-validation trials of the optimal parameter configurations (where the images are divided up and dictionaries $D_1,D_2,\dots$ are used), (2) an example case where the test image can be identified from the training class labels, and (3) an example case where the test image is anomalous.}\label{alg:anomaly}
\end{figure*}

Notice that once a model is trained, the bulk of the work is simply classifying several patches from the test image and utilizing the ensemble classifier (as well as checking whether it is an anomaly or not as we discuss later).  Table \ref{tab:cost} shows the processing times required for a single patch and the entire set of patches associated with a target, when implemented on a 64-bit Core 2 Duo PC with 8 GB of RAM. From this we can see that for any reasonable computer, model design and classification are computationally feasible. Depending on the operator's computational resources, real-time implementation is also conceivable due to the highly parallelizable nature  of the computations involved.

To tune the parameters for PCS, specifically the sparsity values for each class, we used simple cross-validation trials with a training set.  Recall from equation \eqref{eq:lnbayesk} that our spike and slab framework condenses down to solving for $\alpha$ and $\{\xi_k\}_{k=1}^K$.  While $\alpha$ can be set to an appropriately small value, the class-specific terms $\xi_k$ that control sparsity are the challenging aspect that deserve some finer tuning.  In practice, we found that the larger the patch dimensions, the less sensitive PCS was to perturbations to parameters, making them much easier to tune and less prone to issues later on in testing.  The analysis window for which $\xi_k$ can be taken from depends on the application.  For our work, we found values within $(0\,10^{-4}]$ to suffice when given between twenty five and thirty trials.

\section{Anomaly Detection}\label{sec:anomaly}
Anomalous objects are those that are foreign to the training set.  Being able to identify anomalies in Sonar images can be extremely valuable, if not an outright necessity for a fully capable Sonar ATR algorithm.  If a classifier does not have a system to flag potential foreign objects, then an operator can face either a case where a threatening objects goes by without due respect or one in which a series poorly arranged rocks ends up setting off alarms.

Anomaly detection has been widely studied in image processing and machine learning \cite{chandola2009anomaly,pimentel2014review}.  There are several different ways to approach the problem but none specifically tailored for sparsely constrained models like our own.  We propose a novel method that utilizes statistics already calculated for the PCS classification procedure to capably differentiate between an in-training sample and an anomaly.

\begin{figure}[t]\centering
\begin{subfigure}[t]{.48\columnwidth}\centering
\includegraphics[width=.9\columnwidth]{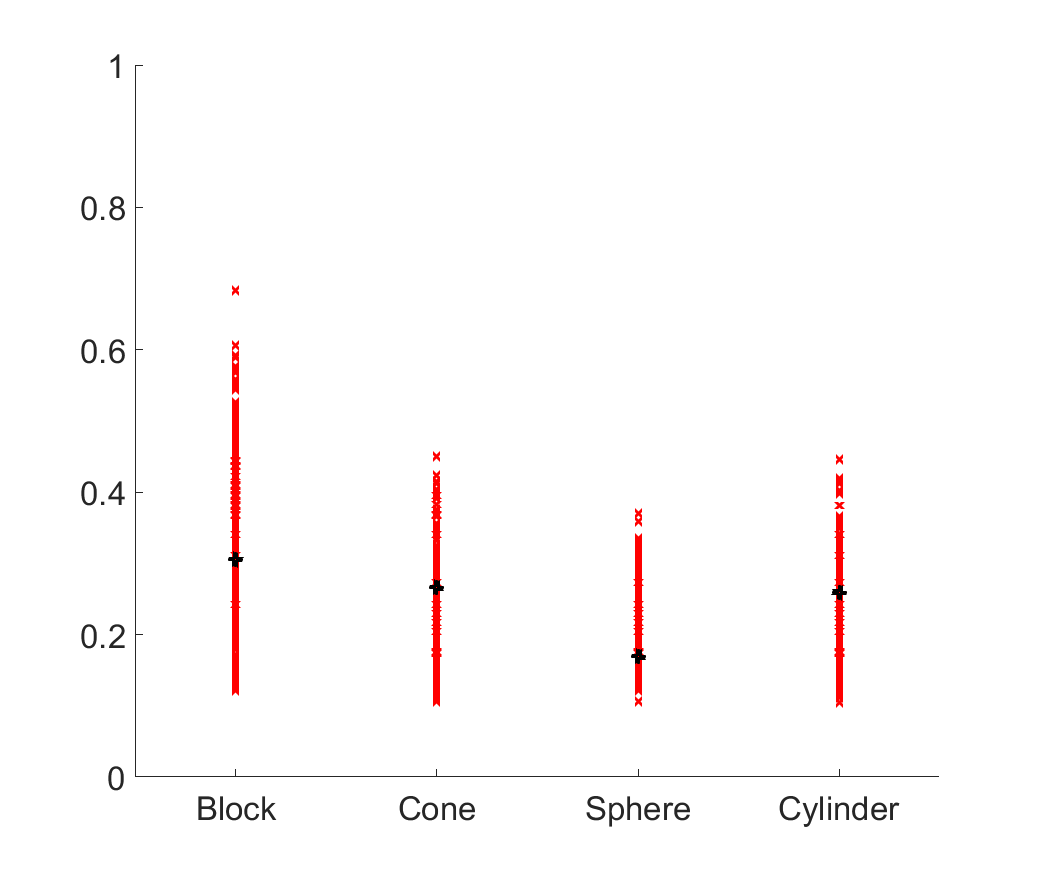}
\caption{Distribution of likelihood values of block patches.}\label{fig:cvdist}
\end{subfigure}
\begin{subfigure}[t]{.48\columnwidth}\centering
\includegraphics[width=.99\columnwidth]{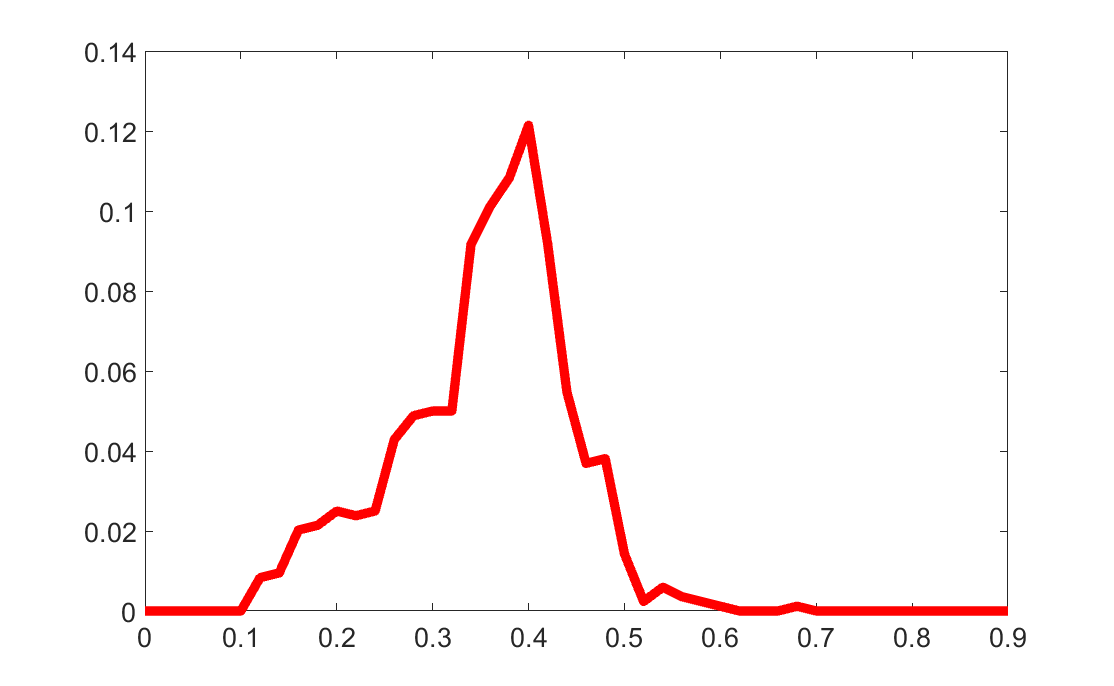}
\caption{Frequency of values from the block class of \ref{fig:cvdist}.}\label{fig:blockcv}
\end{subfigure}
\caption{Example results of the cross-validation stage that are used for the reference distributions for anomaly detection.  Once a suitable set of parameter values are determined after several cross-validation trials, one of the classes could have produced $K$ (4 in this case) vectors of likelihood values like those seen in \ref{fig:cvdist} since each patch yields likelihood value for each class.  An example of a single patch's results are shown by the black marks.  Since we are only concerned witht he distribution associated with the actual class - block - then we single that one out and create a frequency map like that in \ref{fig:cvdist}.}\label{fig:cvanomaly}
\end{figure}

Recall that our PCS algorithm requires a cross-validation phase to find suitable parameters for subsequent classification attempts.  In this, the totality of the patches associated with a single class generates a series of distributions that like shown in \ref{fig:cvdist} where a PCS model was trained on 4 mine-like object classes (blocks, cones, spheres and cylinders - examples of these can be found in Section \ref{sec:experiments}).  While every patch produces $K$ values (a classification yields a value for each class), we are only going to be concerned with the one associated with the known class.  That is, given that the distributions of \ref{fig:cvdist} were from the patches of blocks, then we isolate down to just the blocks vector and try to get a distribution like that of \ref{fig:blockcv}.  This \emph{reference distribution}, which is constructed with the scaled frequency of values within finely-sized bins, serves as our comparison tool to determine an anomaly.

Anomalous determinations are made of test images by similarly taking the patch-level likelihood data.  Here we isolate the distribution associated with whichever class PCS assigned to the image and construct the frequency distribution as we did with the cross-validation data.  Once equipped with these two distributions, the sensible thing would be to compare the two.  If they are close enough, then we can assume that the test image is indeed at least something in our training set.  If not, then some flag would be warranted to notify an operator that an anomaly has been found.  To make such a determination of ``closeness,'' we use the two-sample Kolmogorov-Smirnov (KS) test.  This well known statistical measure for measuring the difference between two distributions is given by
\beqa\label{eq:kstest}
D_{H,I}=\sup\limits_x[F_{1,H}(x)-F_{2,I}(x)]
\eeqa
where $F_{i,J}$ is the empirical cumulative distribution of the $i^\text{th}$ data set (which is the frequencies of the references and tests in our case) containing $J$ (denoted by $I$ and $H$) number of samples.  The pair $F_{1,H}$ and $F_{2,I}$ is deemed significantly different if 
\beqa\label{eq:ksconfidence}
D_{H,I}>c(\alpha)\sqrt{\tfrac{H+I}{HI}}
\eeqa
where $c(\alpha)$ is the value of the inverse of the Kolmogorov distribution at $\alpha$ \cite{darling1957kolmogorov}.  The term $\alpha$ plays a confidence role similar to chi-squared tests, so setting it to the right value comes with certain frequentest caveats, though we use it more as a threshold.

\begin{figure}[t]\centering
\begin{subfigure}[t]{1\columnwidth}\centering
\includegraphics[width=.49\columnwidth]{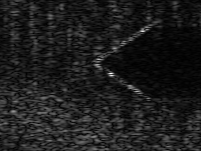}
\includegraphics[width=.49\columnwidth]{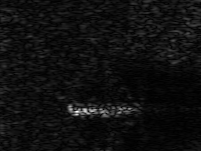}\\\vspace{.1cm}
\includegraphics[width=.49\columnwidth]{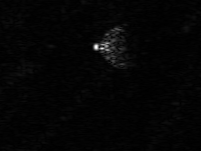}
\includegraphics[width=.49\columnwidth]{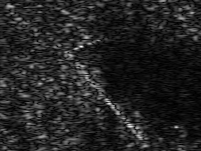}
\caption{Block, cone, sphere, and cylinder mines making up our SAS dataset in that order.}\label{fig:targets}
\end{subfigure}
\begin{subfigure}[t]{1\columnwidth}\centering
\fbox{\includegraphics[width=.99\columnwidth]{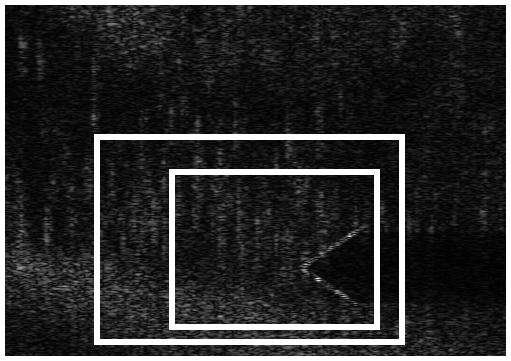}}
\caption{Example window sizes of expansive, middling, and narrow.}\label{fig:targetWindows}
\end{subfigure}
\caption{SAS images used in Sonar ATR experiments.}\label{fig:data}
\end{figure}

The entire procedure is summarized in Figure \ref{alg:anomaly}.  Note in our illustration we included a bimodal frequency distribution which, in practice, was not uncommon.  Reference distributions took on a fairly ranging series of shapes.  As well, while we present our anomaly algorithm specifically for PCS, any method that utilizes our patch classifier strategy could use our proposed idea. 

\section{Experiments}\label{sec:experiments}
To illustrate the advantages of PCS, we show the following results using a Sonar image dataset provided by the U.S. Naval Surface Warfare Center.  This collection of Sonar images contains five mine-like objects, blocks, cones, cylinders, spheres, and toruses, with considerable variability with regards to the targets' positions and background patterns.  For the blocks, cones, cylinders, and spheres we had sixty samples each while only twenty two of the toruses.  Since blocks and cylinders can look vastly different based on the capture position, the angles were varied from an initial set up between fifteen to seventy five degrees for the blocks and zero and one hundred twenty for the cylinders.  The backgrounds for each image come from actual SAS images captured on UAVs and the objects are simulated upon them using proprietary Naval software.  

For each class of objects we used three different capture sizes to represent an increasing difficulty and uncertainty for the desired target.  That is, we have three test sets:  a narrowly captured group that has the least amount of background clutter to interfere with a classification, a middling set where the target has more room for pose variability within the chip and some clutter, and lastly an expansive set wherein a great deal of background can present itself to confuse a classifier.  Given the importance of reliability with regards to a Sonar ATR algorithm, the ability to perform well on any of these three types of images - with special weight on the realistically difficult expansive set - is imperative for its feasibility in real-world settings.  Figure \ref{fig:targets} offers examples to both the classes of objects as well as the three capture scenarios note SAS images shown in previous sections are also useful examples.

In our first experiment we show a comparison between our PCS method using the spike and slab Bayesian framework and a similar strategy using an $\ell_1$ solver to illustrate the necessity of our more probabilistic approach to our SAS classification problem.  Next, we demonstrate the performance of our method against existing Sonar ATR algorithms over varied training sizes to not only show the quality of PCS in the context of other options but also to demonstrate how well it thrives in limited training settings.  Lastly, we cover the impact of Rayleigh noise upon our Sonar images and how PCS is able to handle such an imperfection even in the more intense cases.  Note that each model was trained using images from the narrow set as these offered the most object-specific features and little background interference.

\subsection{Assessment of Spike and Slab for PCS}\label{subsec:pcs}
\begin{figure}[t]\centering
\begin{subfigure}[t]{1\columnwidth}\centering
\begin{tabular}{c|c|c|c|c|}
& (B)locks & (Co)nes & (S)pheres & (Cy)linder\\\hline
Trial 1& {\color{violet}B} {\color{blue} Co} &  {\color{violet}B} {\color{blue} Co}&  {\color{violet}S} {\color{blue} S} &  {\color{violet}S} {\color{blue} Cy}\\
Trial 2&  {\color{violet}B} {\color{blue} B} & {\color{violet}S} {\color{blue} Co} &  {\color{violet}S} {\color{blue} S} &  {\color{violet}S} {\color{blue} Cy}\\\hline
\end{tabular}
\caption{Results of eight tests of the exact same images and patches for PCS with L1LS ({\color{violet}violet}) and spike and slab ({\color{blue}blue}).}\label{tab:ssll}
\end{subfigure}
\begin{subfigure}[t]{1\columnwidth}
\includegraphics[width=.32\columnwidth]{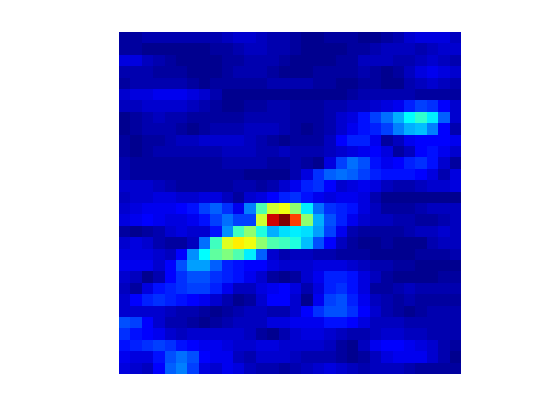}
\includegraphics[width=.32\columnwidth]{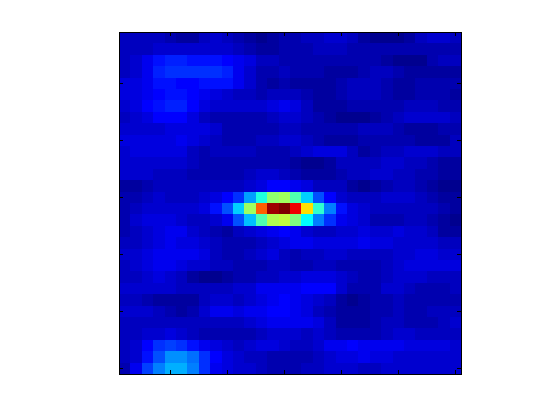}
\includegraphics[width=.32\columnwidth]{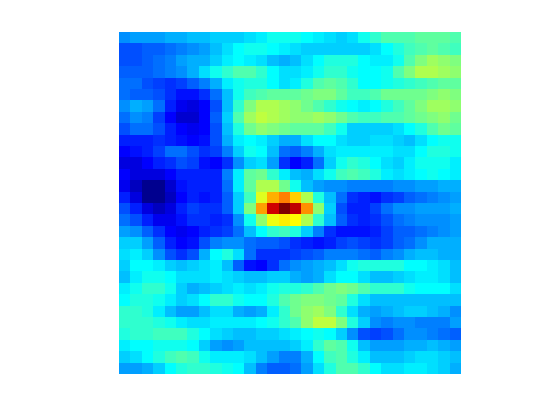}
\caption{Example patch reconstructions.  Left is original, middle is L1LS, and right is spike and slab.}\label{fig:reconstructs}
\end{subfigure}
\begin{subfigure}[t]{1\columnwidth}
\includegraphics[width=.47\columnwidth]{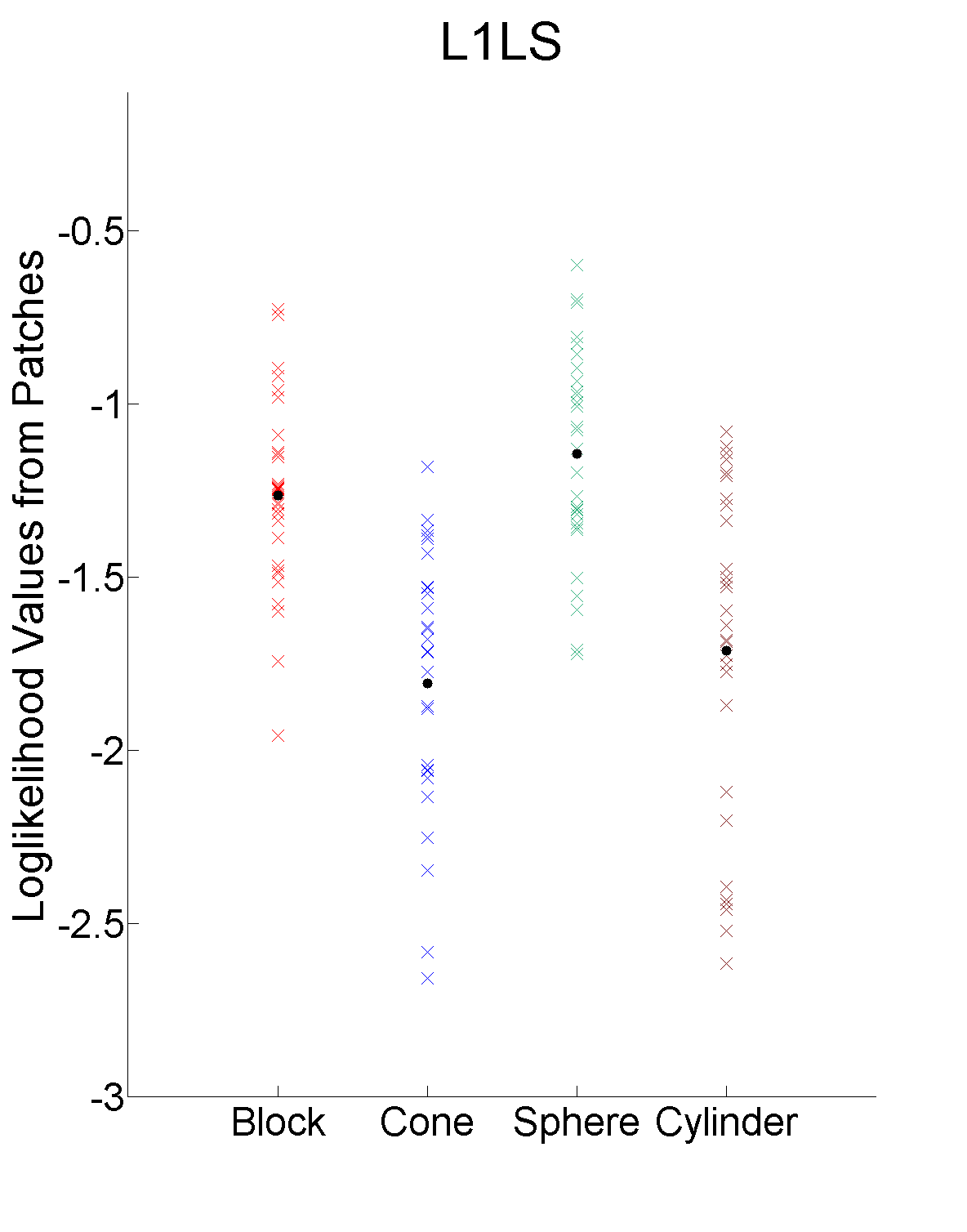}
\includegraphics[width=.47\columnwidth]{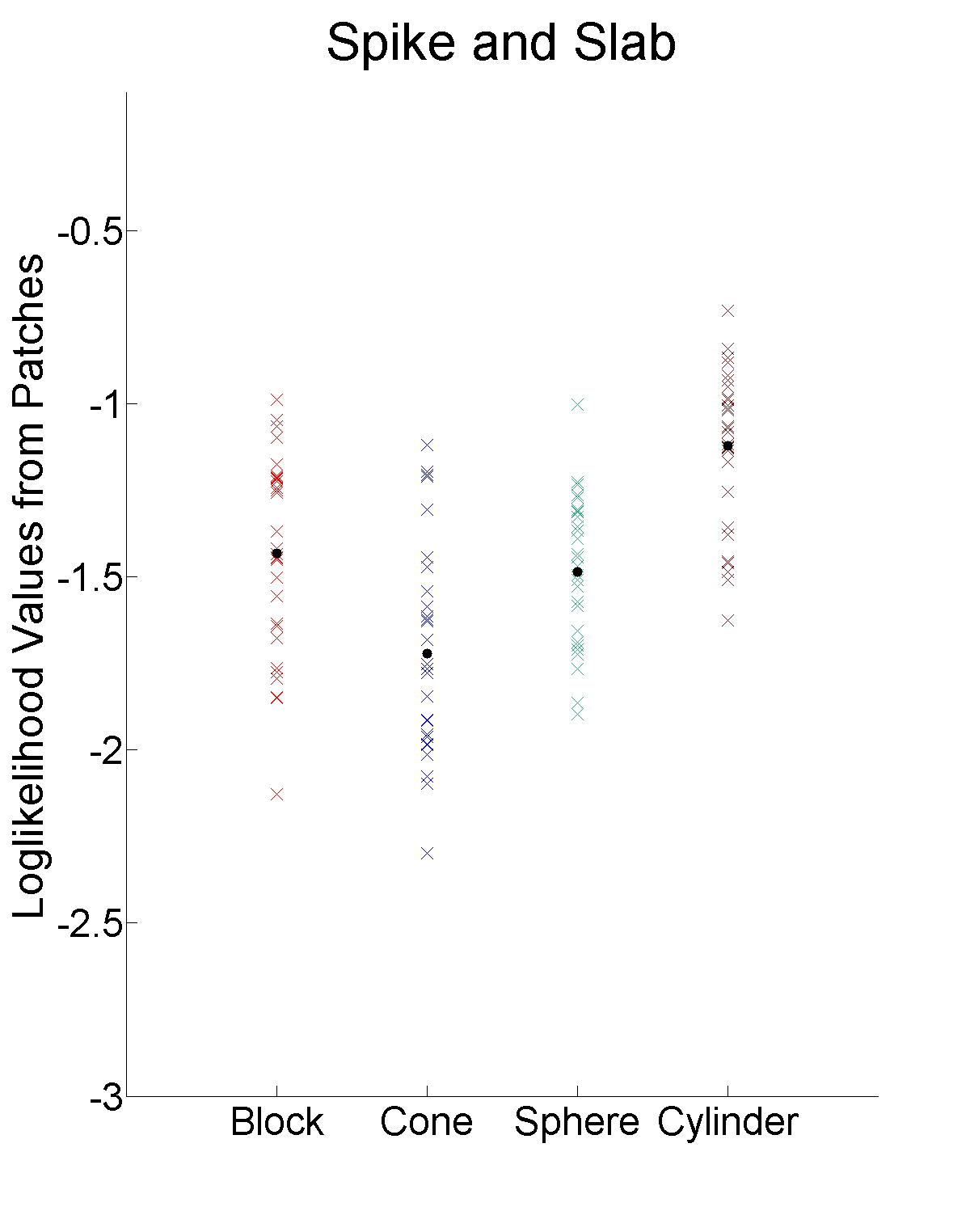}
\caption{Loglikelihood values corresponding to the cylinder test of the second trial.  The left is the L1LS modified PCS and the right is the unaltered PCS with spike and slab.  The black dots refer to the means.}\label{fig:sslldistributions}
\end{subfigure}
\caption{PCS with spike and slab vs. PCS with L1LS.}\label{fig:pcsl1ls}
\end{figure}

One of the most important aspects to our PCS algorithm is the spike and slab prior model for handling class-specific priors.  As we described, it contains a level of distinction lacking in current popular $\ell_1$ approaches.  The idea is that Sonar images lack many of the discriminating features common in other classification problems and an greater sense of separation between classes is necessary for a sufficient classifier.  Here, we pick a known and widely used $\ell_1$ relaxation technique, L1LS \cite{kim2007interior} to compare our spike and slab construction against and did so by simplying taking the framework of the PCS algorithm and swapping out our Bayesian prior with L1LS.  As seen in \ref{tab:ssll} shows the results on eight test images (two per class and both from the middling set) along with an example patch reconstruction, its associated coefficient plots, and resulting likelihood values.

What is most interesting about our spike and slab approach as opposed to the traditional SRC is that we are \emph{not} looking to for the best reconstruction of the test patch.  The theory behind SRC is based upon reconstruction error and seeks to pick the sparsest solution that best limits this value.  The spike and slab approach does not do this.  By creating different tiers of potential sparsity, we \emph{limit} the ability of the model to best reconstruct the image - but this \emph{does not matter}!  Our classification problem is not centered around rebuilding the image but ensuring that the right class yields the lowest relative residual error even if that means raising the total error.  In this regard, the spike and slab construction yields an advancement in the root ideas underpinning sparsity-based classification.  The results of Figure \ref{fig:pcsl1ls} demonstrate this point; overall, under the exact same circumstances, i.e the same dictionary, same test images, and even the same extracted patches, spike and slab was able to well outperform L1LS which got stuck between only classifying objects as spheres or blocks.  This came, though, with excellent image reconstructions by L1LS such as those illustrated by patches \ref{fig:reconstructs}, especially when compared to the attempt made by the spike and slab model.  This factor did not matter.  In almost every instance we ended up with a distribution of loglikelihood values similar to that shown in \ref{fig:sslldistributions}:  the spike and slab values tend to favor the actual class while the L1LS teeter between blocks and spheres.

\begin{figure*}[t]\centering
\includegraphics[width=.325\textwidth]{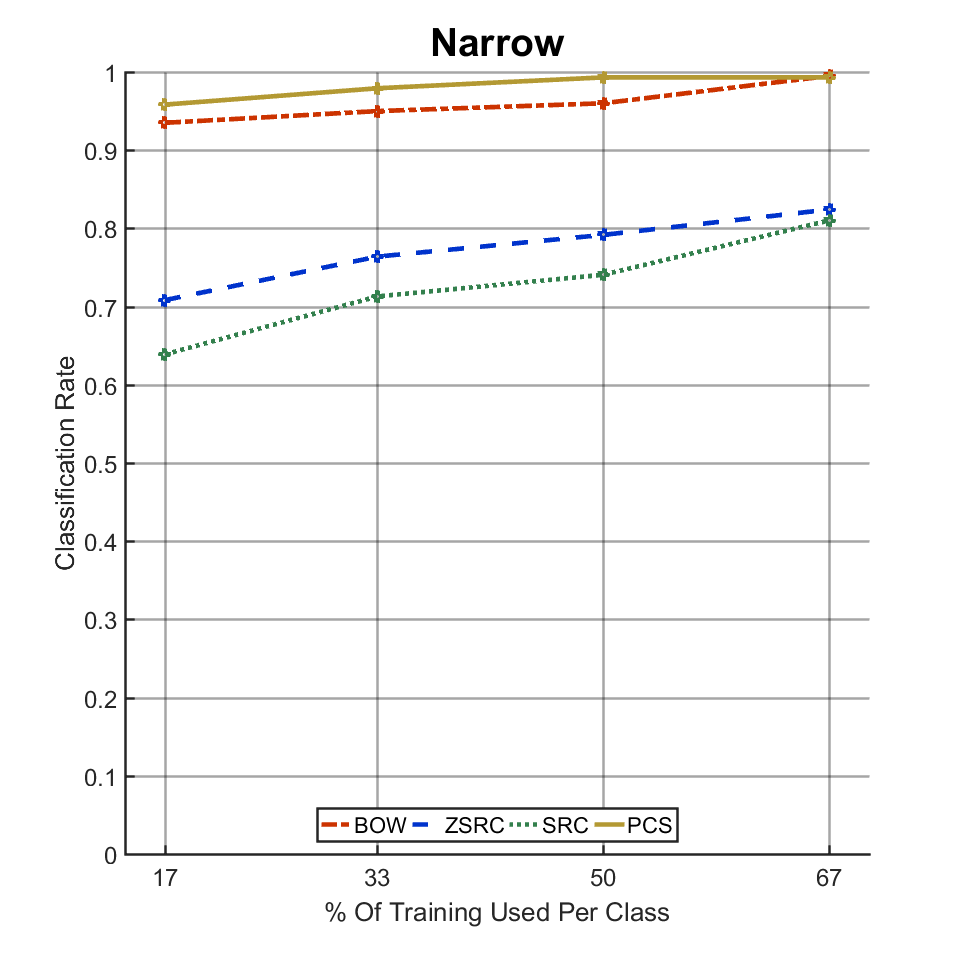}
\includegraphics[width=.325\textwidth]{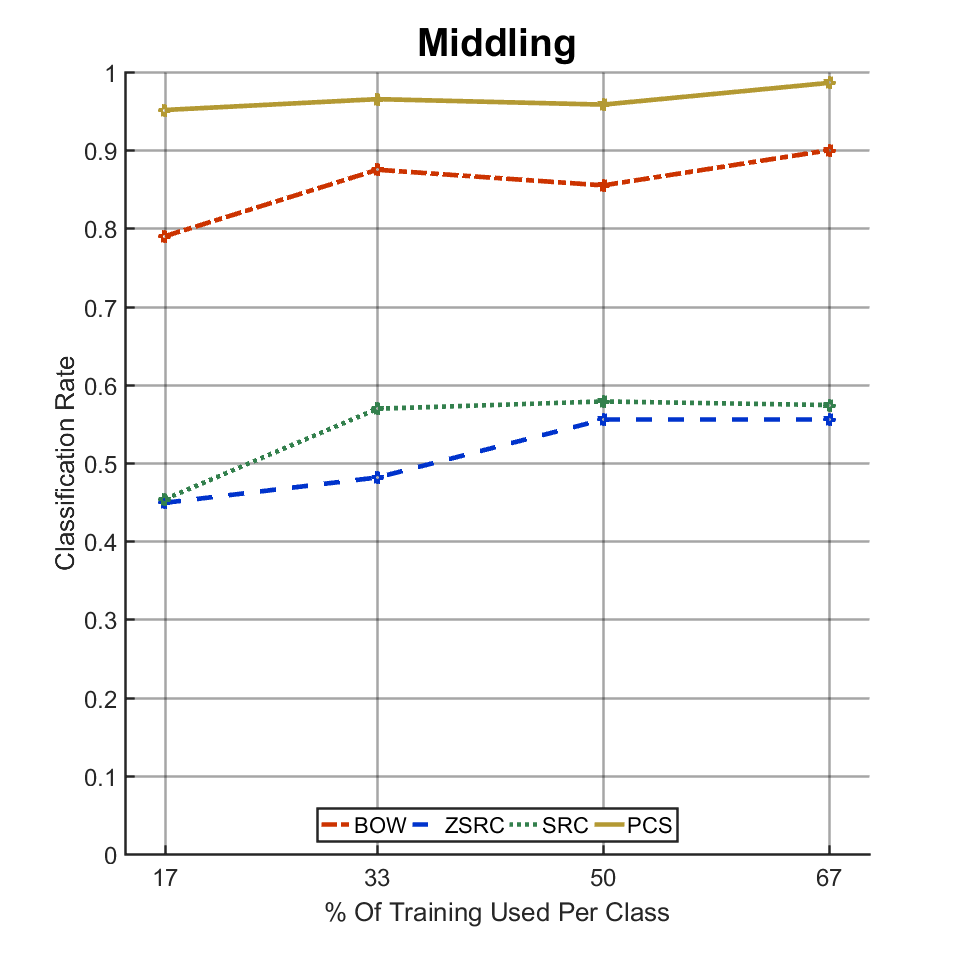}
\includegraphics[width=.325\textwidth]{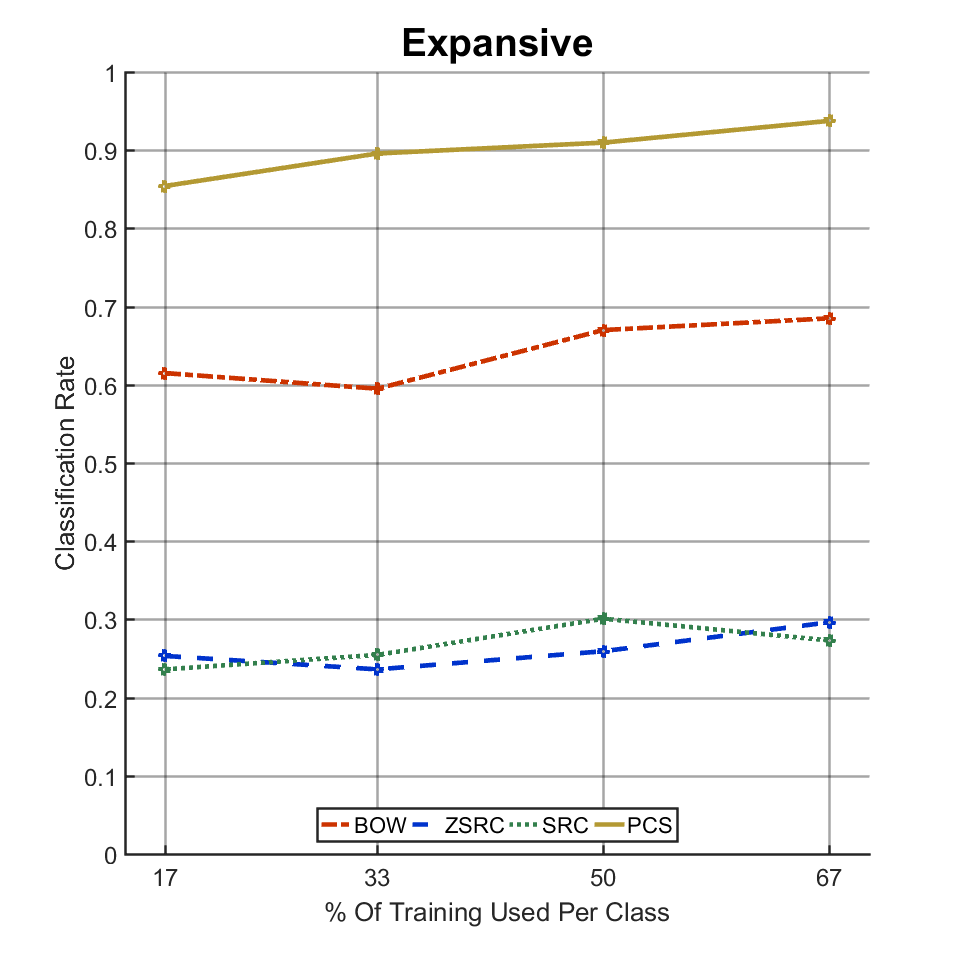}
\caption{Classification rates across varied training sizes for the three window sizes.  There were a total of sixty images per class and we ranged training from ten to forty.  {The mean $\xi_k$ values for PCS across tests was $\bol{\xi}^T=[.001,.002,.012,1.052]\times 10^{-4}$.}}\label{fig:sizeComp}
\end{figure*}

\begin{figure*}[t]\centering
\includegraphics[width=.325\textwidth]{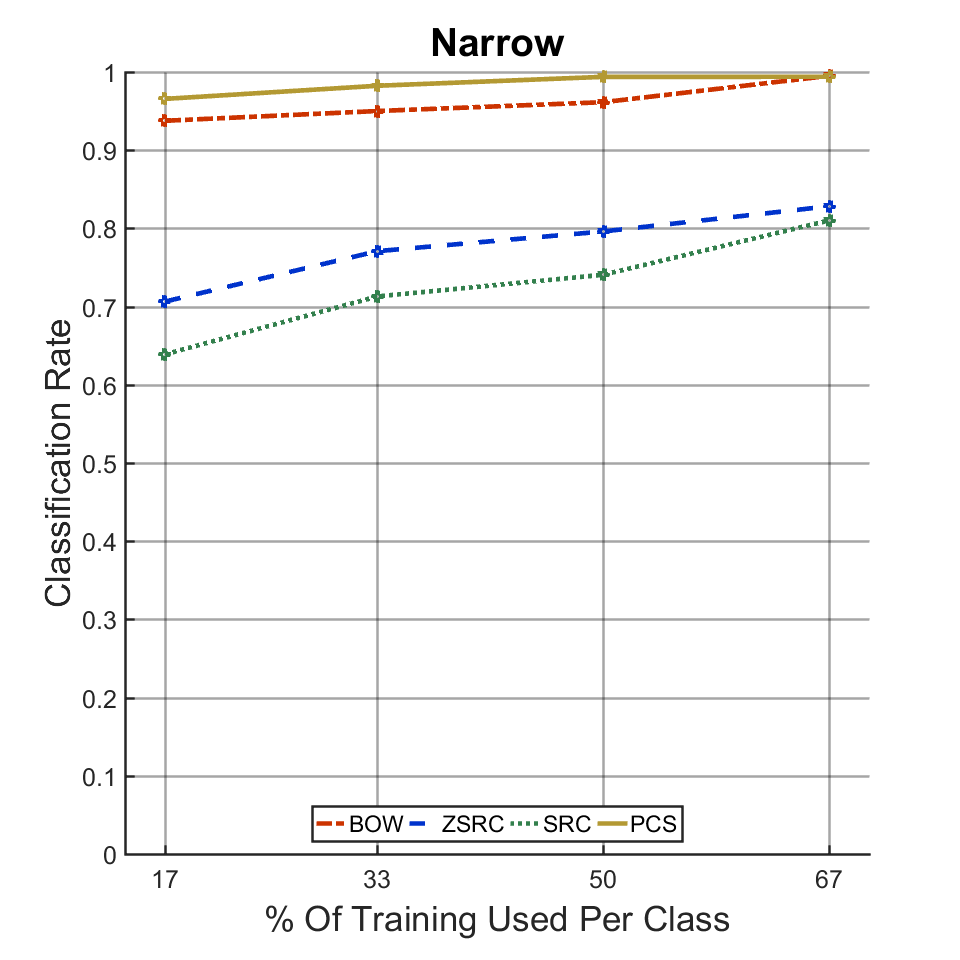}
\includegraphics[width=.325\textwidth]{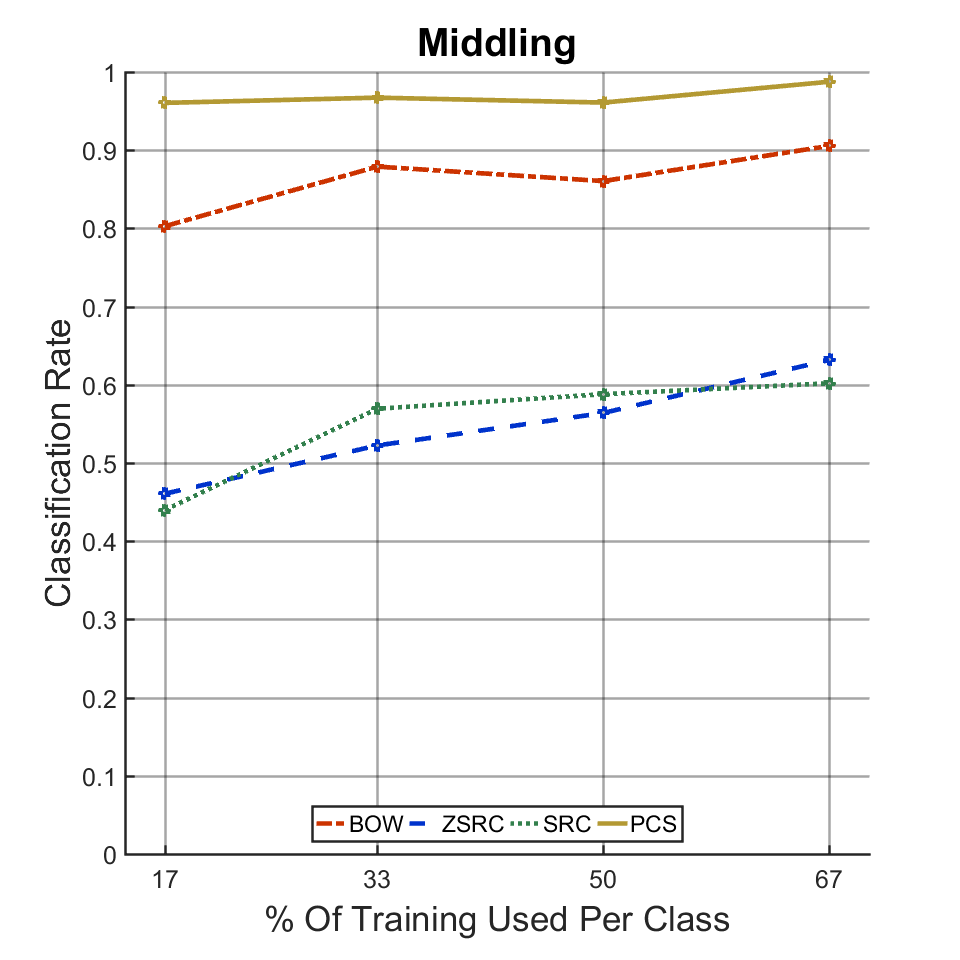}
\includegraphics[width=.325\textwidth]{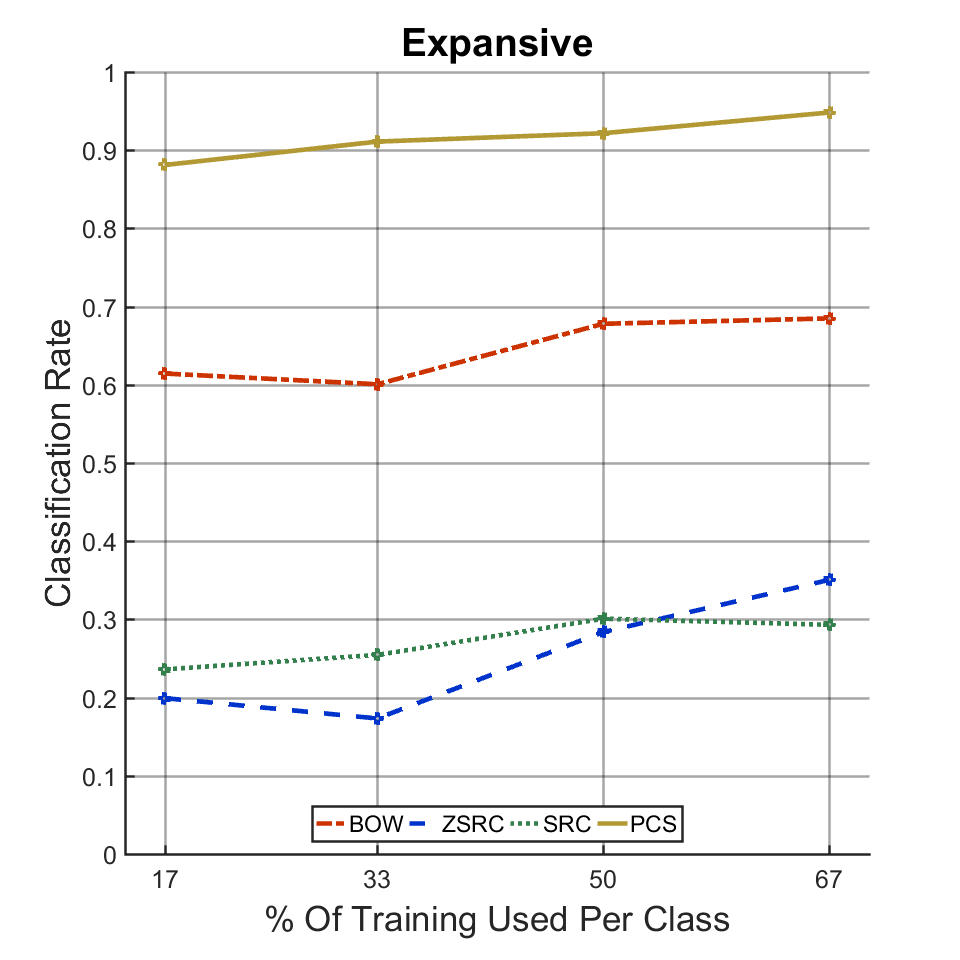}
\caption{Mean precision values for 4 class Sonar problem with varied training.  Plot titles indicate window size for tests.}\label{fig:sizep}
\end{figure*}

\subsection{Comparisons against well-known sonar ATR methods}

We look to address the relative strength of PCS compared to that of several other powerful classifiers.  For the following, we compare against three state of the art methods:
\enum[$\circ$]
\ii \textbf{Zernicke Moment-Based SRC (ZSRC)} \cite{kumar2012object}:  this approach utilizes Zernicke Moments as transformations to address geometric variability in Sonar targets.  The authors ares able to illustrate how by using a SRC framework with Zernicke Moments instead of images themselves, compelling mine identification rates can be achieved in a straightforward manner.  For us this serves as a useful comparison of modern sparsity-based methods in Sonar ATR and how their feature-based approach differs in execution.
\ii \textbf{SIFT Bag of Words (BOW)} \cite{zhu2014model}:  SIFT feature bag-of-words models are very well known in computer vision circles and serve as a familiar approach for many readers.  The authors used such a classifier to identify targets in simulated Sonar settings.  The training for this model ended up being the most sensitive to background clutter and, while we do not show it here, using the larger windowed image sets for training impacted this method substantially more than any other.
\ii \textbf{Sparse Reconstruction-Based Classification (SRC)} \cite{fandos2009sparse}:  Since we have proposed a more involved sparsity model built off of the ideas underpinning SRC, it makes sense to incorporate it in our experiments.  The authors here performed a simple test of SRC upon Sonar targets and showed it to be a viable option, though they did not discuss ways to creatively handle different target pose issues.
\eenum
With these, we focus on two applications:  limited training and noisy image classifications.  Note that for all the limited training trials, six different partitions of training and testing data were made for each scenario and every algorithm used the same data arrangements.  Thus, for the case where 17\% of the available images were used for training, we had six models per algorithm and comparisons could be made without the concern of one method having a favorable training/testing.  When it came to noise, we used the six models per algorithm resulting from the highest training scenario.  In every case, nine test images were used per class, meaning thirty six in total.

\begin{figure*}[t]\centering
\includegraphics[width=.325\textwidth]{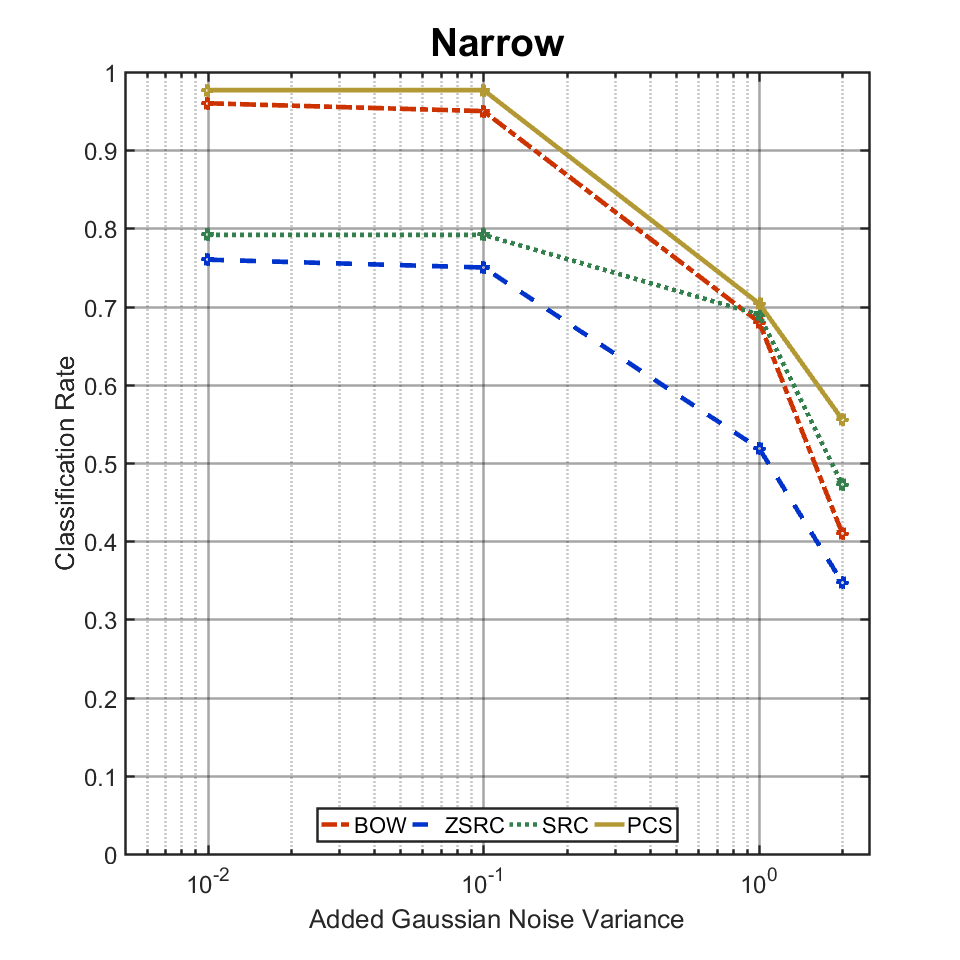}
\includegraphics[width=.325\textwidth]{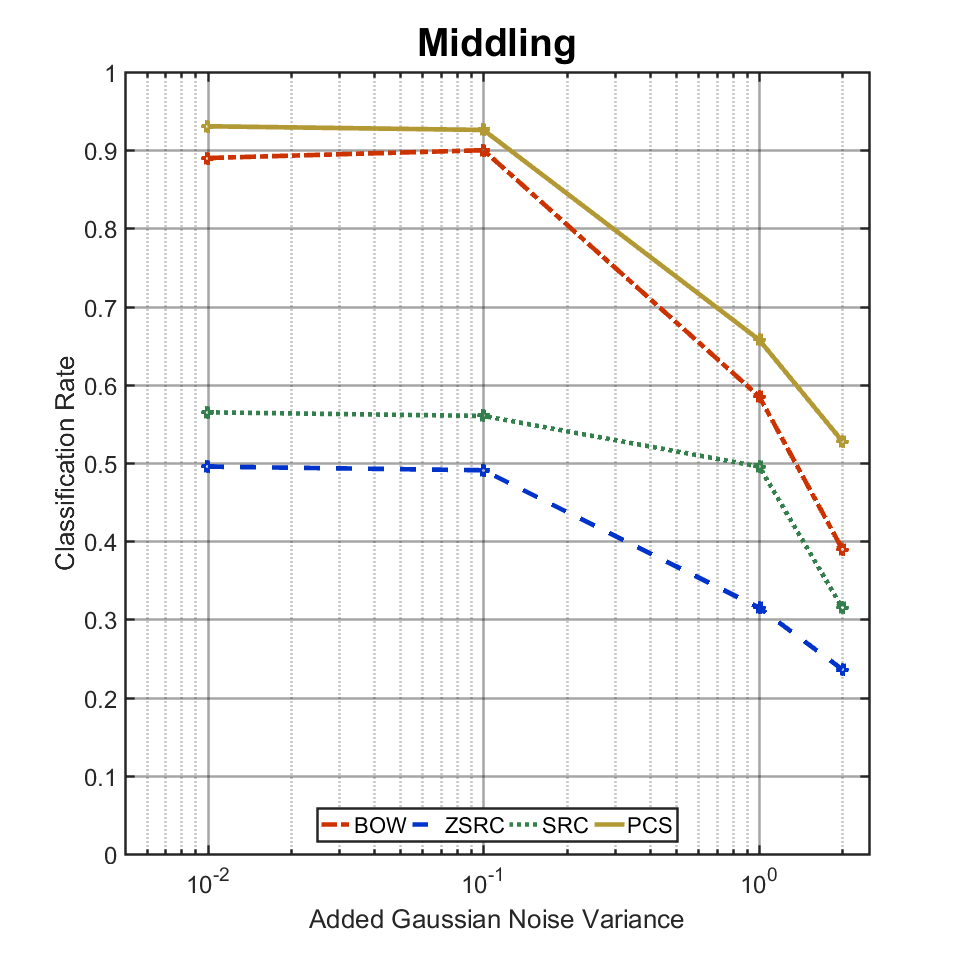}
\includegraphics[width=.325\textwidth]{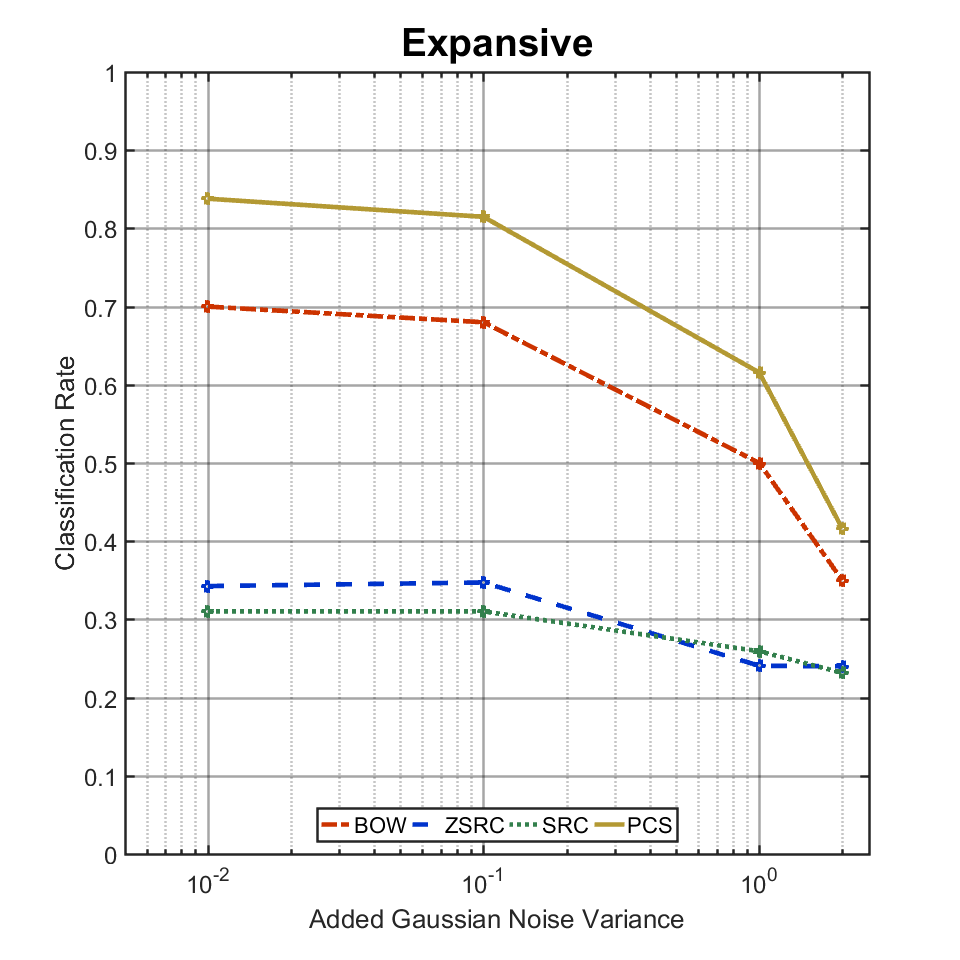}
\caption{Mean classification rates across increasingly intense Rayleigh noisy images.}\label{fig:noiser}
\end{figure*}

\begin{figure*}[t]\centering
\includegraphics[width=.325\textwidth]{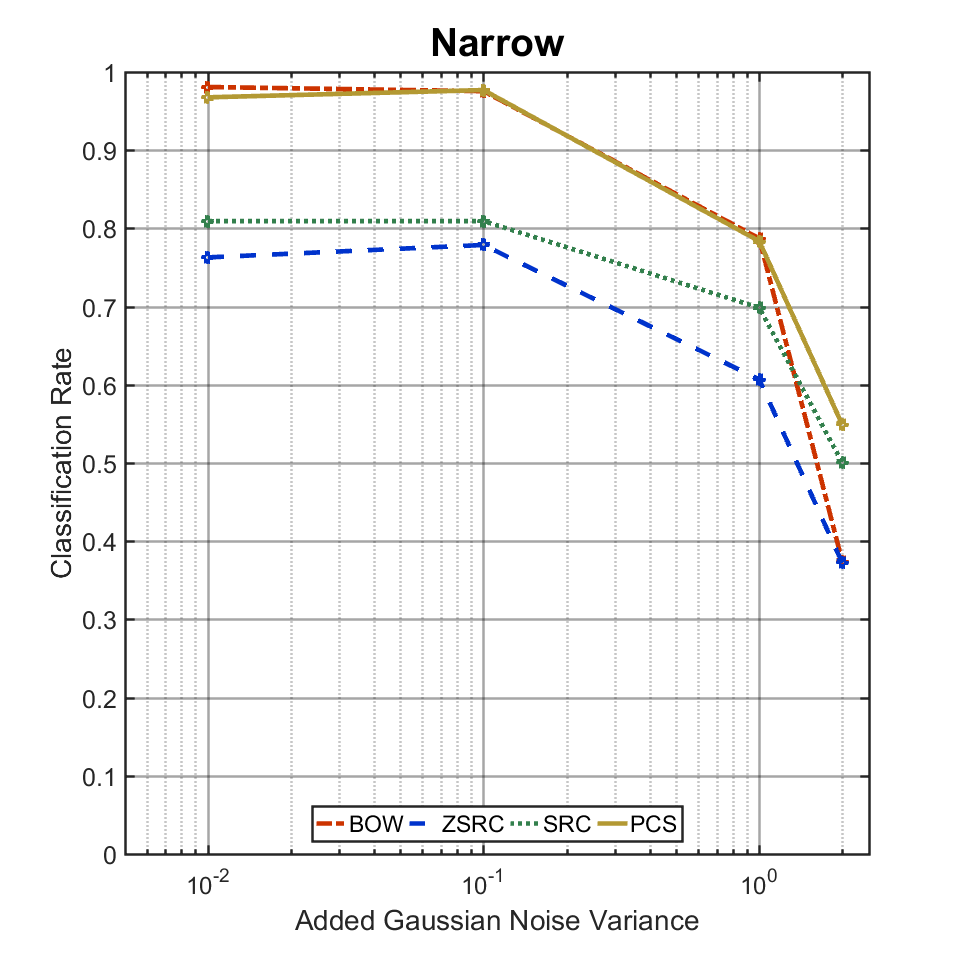}
\includegraphics[width=.325\textwidth]{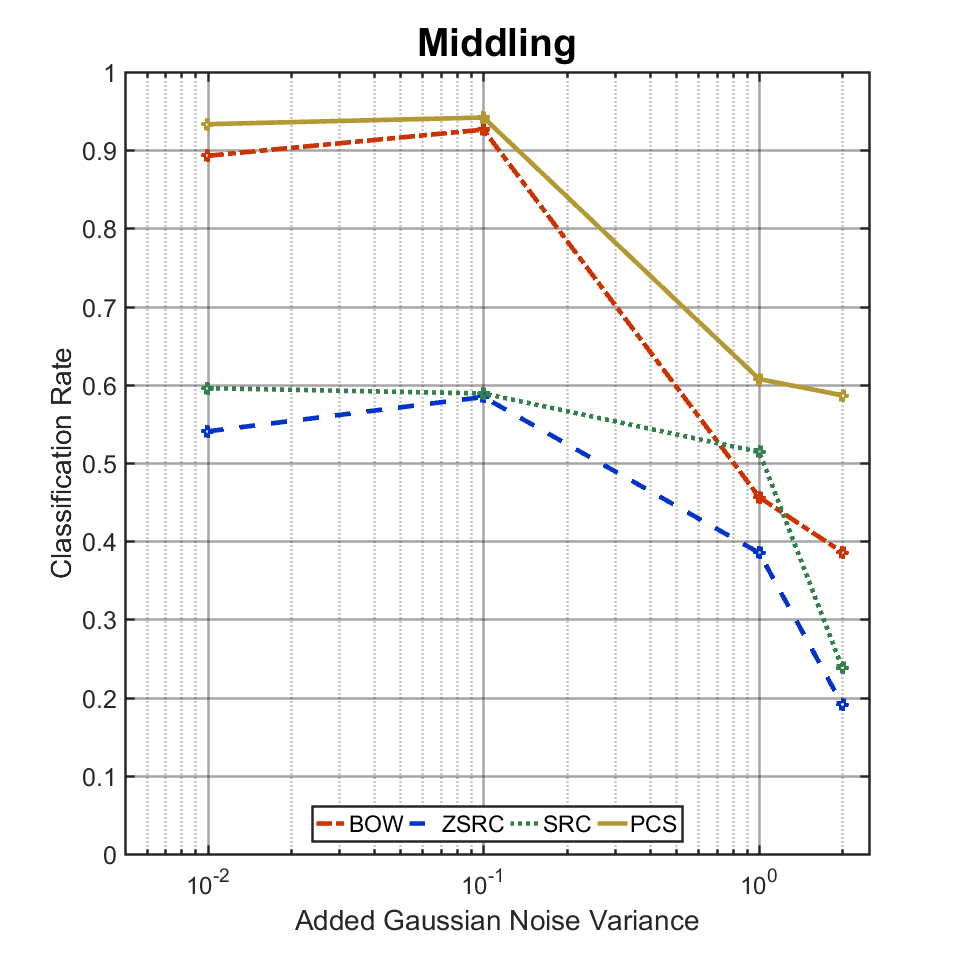}
\includegraphics[width=.325\textwidth]{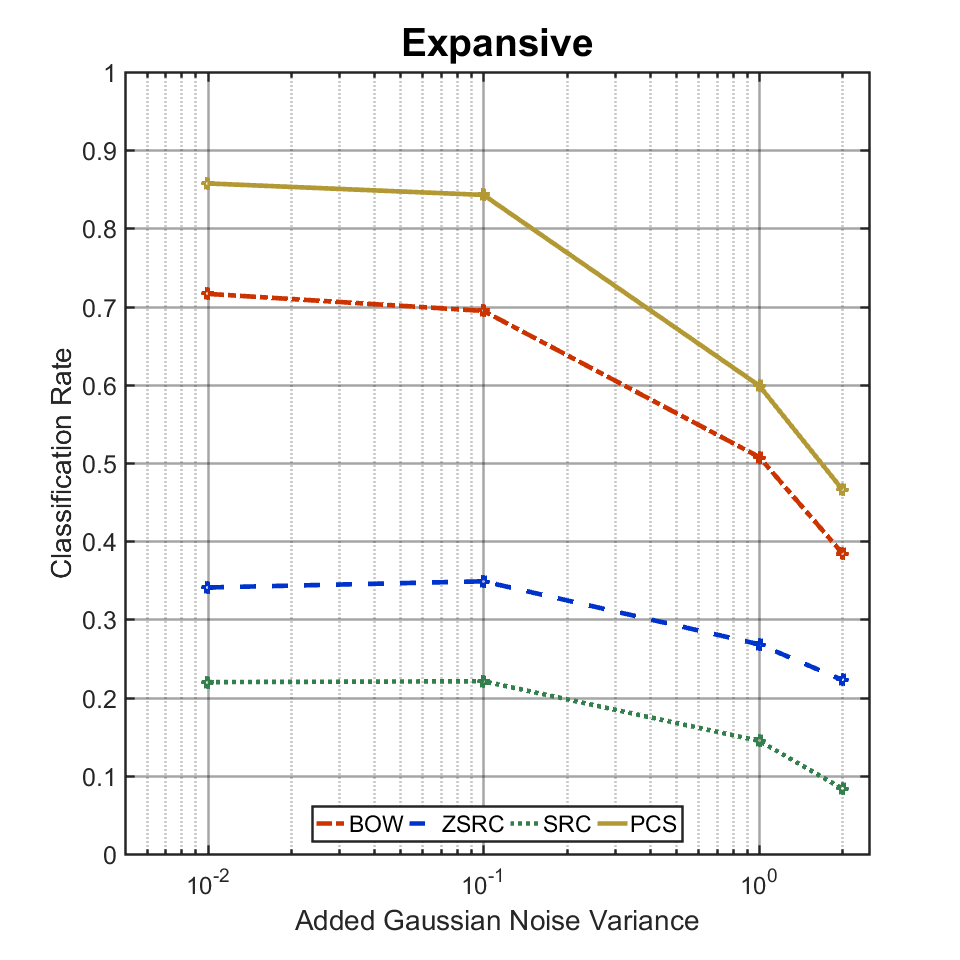}
\caption{Mean prevision values across increasingly intense Rayleigh noisy images.}\label{fig:noisep}
\end{figure*}

For the limited training setting, it is important to realize that Sonar images can be difficult to come by.  AUV treks can be expensive and timely to carry out, not to mention that some items may be infrequently seen - all of which leading to few training samples.  How a classifier performs given poor training is an essential question for any Sonar ATR and we look to specifically address that here.  

Recalling that we have sixty images per class, Figure \ref{fig:sizeComp} provides the overall classification rates for models trained from the narrow set with ten images per class (17\%) up to forty (67\%).  Our patch-based approach yields the highest recall rates of any of the methods in every case.  Unsurprisingly, every method does its relative best in the narrow window tests as they are intentionally the most focused.  From there, we see a decline with widening of the capture setting - steepest for SRC and BOW.  Interestingly, PCS is able to achieve greater than eighty percent recall rates even in the lowest of training settings against the most challenging of images.  Given the patch-based nature of PCS, the fact that it does well with regards to limited training is not necessarily unexpected given that a single image contributes several bits of information towards the dictionary, but its consistency despite larger target sizes is encouraging.  This factor speaks to its overall reliability; while every other method faltered when broached with the middling or expansive sets while PCS saw a substantially less impact.

Of course, recall is only part of the story.  In addition to classifying objects, we have to be concerned with the precision statistics for each object class - that is - the ratio of all images classified as, for example, blocks that are actually blocks.  A classifier can have a passable recall rate that disguises confusion between two or more classes that precision statistics can reveal, as is the case later on in the Noise tests with SRC.  Here, we find similar trends as to what we found with the recall results.  Figure \ref{fig:sizep} shows that, again, PCS did best in retaining the fidelity of each class even in the toughest of settings.  Again, this adds to our reliability narrative for PCS as we have evidence to trust an assessment to not be a false alarm.  Such a case is dangerous and costly for most Sonar ATR applications so this characteristic is key for PCS.

\begin{figure}[b]\centering
\includegraphics[width=.235\columnwidth]{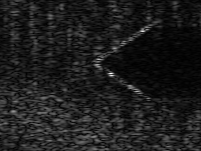}
\includegraphics[width=.235\columnwidth]{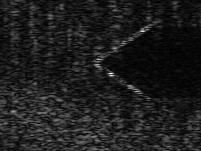}
\includegraphics[width=.235\columnwidth]{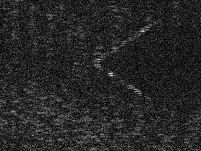}
\includegraphics[width=.235\columnwidth]{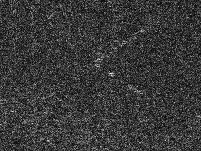}
\caption{Original test image with increasing Rayleigh noise to highest intensity (left to right: no noise ,$\sigma= .1$, $\sigma=1$, $\sigma=2$ where variance is $\sigma$.  Note that each image was normalized).}\label{fig:exampleNoise}
\end{figure}

As for noise, the typical white Gaussian noise is not the best to use in sonar images.  The complex-valued nature to Sonar imagery means a Rayleigh model is more appropriate \cite{hayes2009synthetic} since we use the magnitude of these values in our case. In other words, we suppose additive white Gaussian noise for both the real and complex parts of our signals which becomes multiplicative Rayleigh when we create our images.  The variance of those Gaussian distribution is kept equal for both the real and imaginary parts, meaning that it is the same for the Rayleigh distribution we draw noise from.  To test for robustness against noise, we increase this variance of the Rayleigh distribution to simulate an increasing intensity.  The images of Figure \ref{fig:exampleNoise} demonstrates an example of the progression of noise we test with.   Sparse reconstruction-based classifiers have rarely been tested against Rayleigh noise with some considerations made in SAR ATR and echocardiogram image classification \cite{song2016sparse,guo2015automatic}.

For the tests whose mean recall and precision statistics can be found in Figures \ref{fig:noiser} and \ref{fig:noisep}, respectively, we constructed six models for each algorithm off of forty different training images per class and tested them against thirty six Sonar images (nine per class).  PCS again came out ahead of every other algorithm in every case, whether that be in terms of recall or precision.  In the narrow windowed case, we see that the Bag-of-Words model did comparable with PCS except for the highest stage of noise variance, but the gap between our method and BOW, the second best, became more noticeable as we ramped up the difficulty.  To note, the SRC classifier began confusing every class with cones thus giving it the close to twenty five percent accuracy and dismal precision values - a weakness this method faces when geometric diversity addressed as we have.

What does this mean?  Although a Sonar-ATR does not always face extreme noise conditions--such as illustrated in Figure \ref{fig:exampleNoise}--it is nevertheless not out of the question to occasionally obtain such noisy images.  The ability - without post-processing - to handle such SAS images can be crucial in avoiding problematic images.  If the noisy image is nothing of concern or, conversely, a highly dangerous sea-floor target, our experiments show that a well-trained PCS algorithm can at the least provide a competent assessment.  In the much more common examples where there is some but not incredible bouts of Rayleigh-like noise, we see that such a phenomenon has far less impact on PCS than other options.

\subsection{Anomaly Detection}
\begin{figure}[t]\centering
\includegraphics[width=.7\columnwidth]{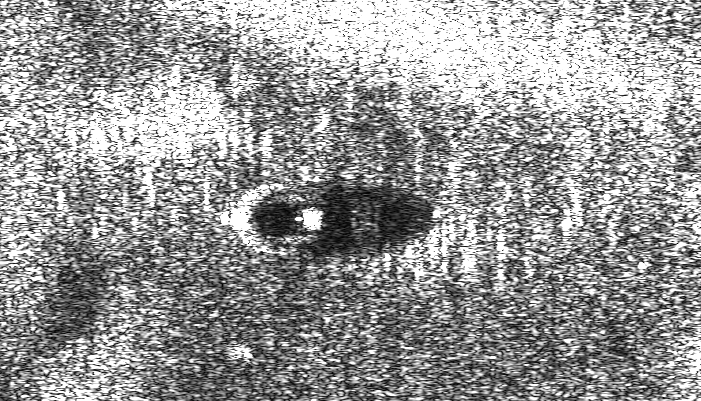}
\caption{Example torus SAS image.}\label{fig:torus}
\end{figure}

Among the data provided in the RAWSAS dataset is a small sample of torus (tire-shaped) SAS images.  Given that there are too few for proper testing (only 22), we find decided to save these for anomaly detection experiments.  Toruses, of which Figure \ref{fig:torus} presents an example, are unique with respect to each of our training classes with a lack of sharp corners and multiple rounded intensity points.  To test our torus images, we used PCS with forty training images per class.  A reference distributions was generated during parameter-setting trials in the manner described in Section \ref{sec:anomaly}.  The confidence value for KS tests was set to $\alpha=.001$.

\begin{figure}[t]
\includegraphics[width=\columnwidth]{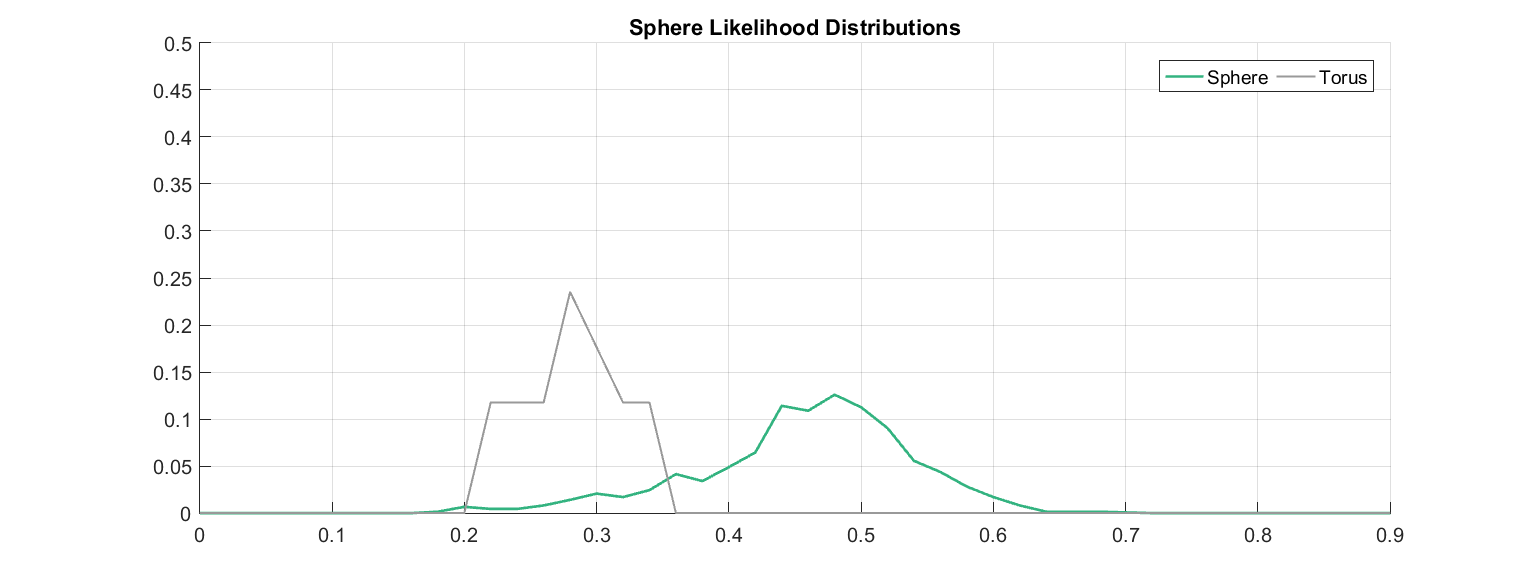}
\includegraphics[width=\columnwidth]{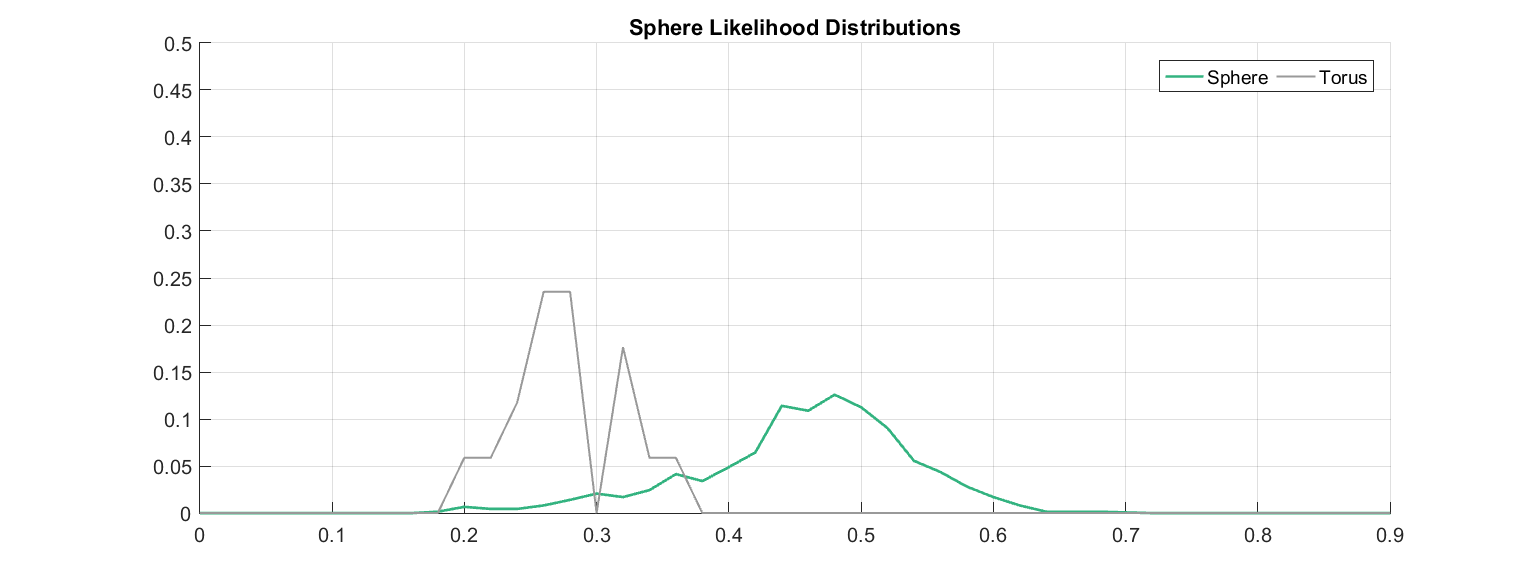}
\includegraphics[width=\columnwidth]{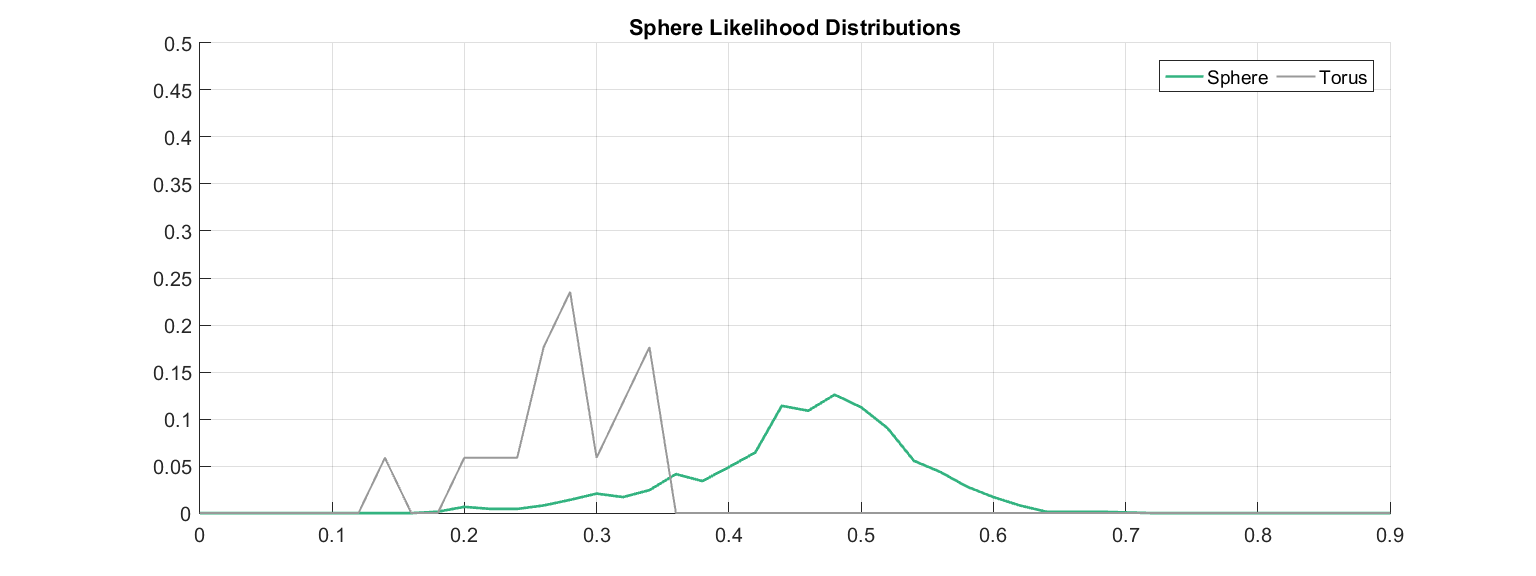}
\caption{Spherical distributions compared against example torus distributions. These three cases were all correctly identified as anomalies.}\label{fig:anomalies}
\end{figure}

Figure \ref{fig:anomalies} depicts the distributions of three of the ten images of which we tested, all of which were flagged as anomalies, and the differences are striking.  The bell-curve-like shape of the reference distribution from the sphere class, the assignment PCS made for every torus, crowds mostly within the range of .4 to .6 while the toruses' centered around the value of .3, reflecting a more confused classification.  The two distributions in each case have clear, visible distinct patterns per their PCS results which allow for our anomaly framework to work.  

\begin{figure}[t]
\includegraphics[width=\columnwidth]{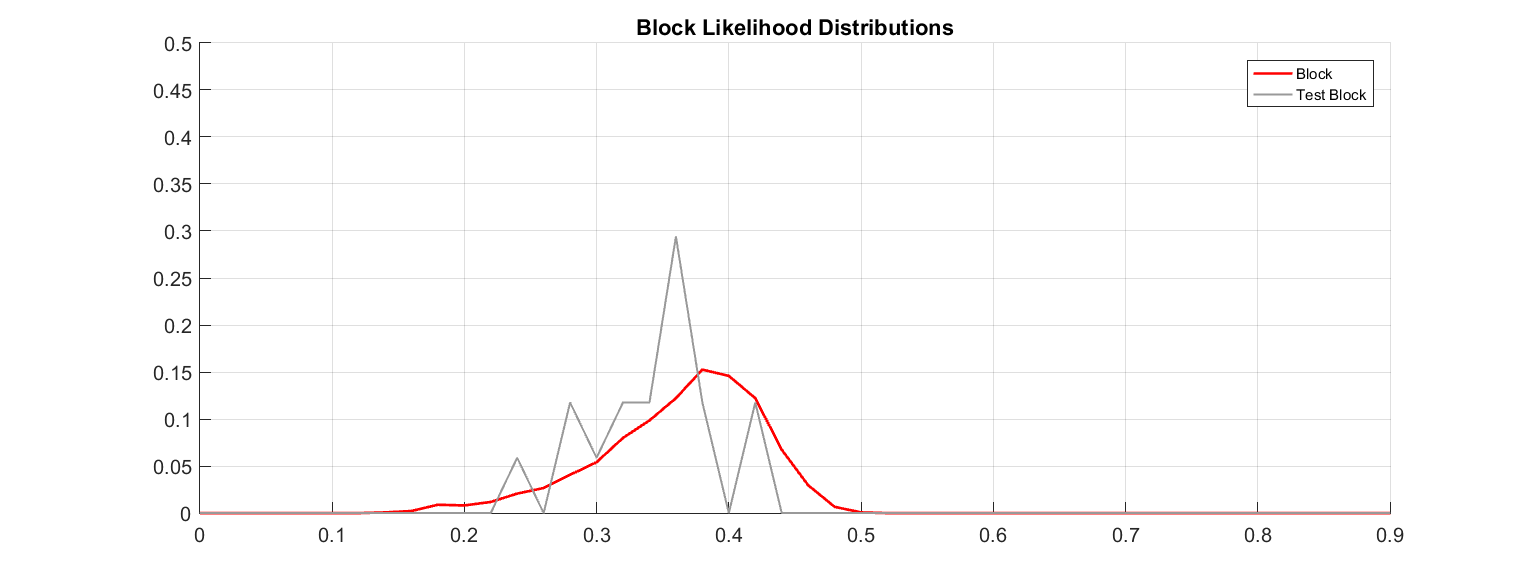}
\includegraphics[width=\columnwidth]{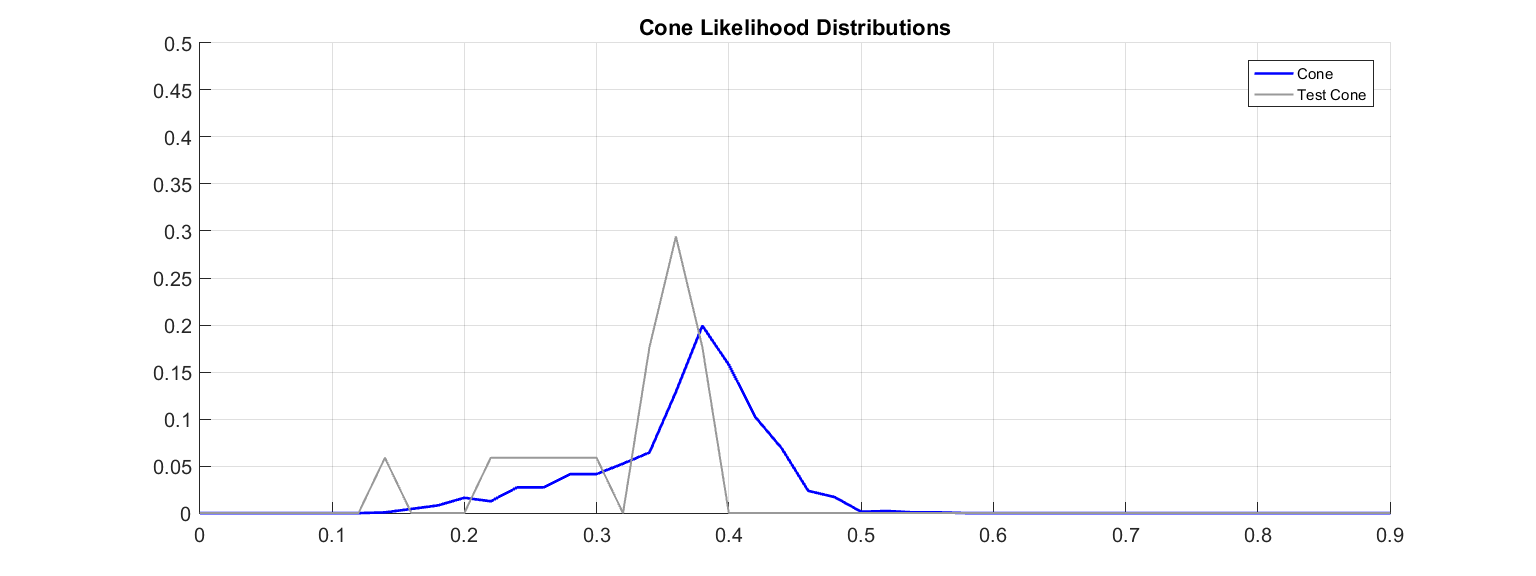}
\includegraphics[width=\columnwidth]{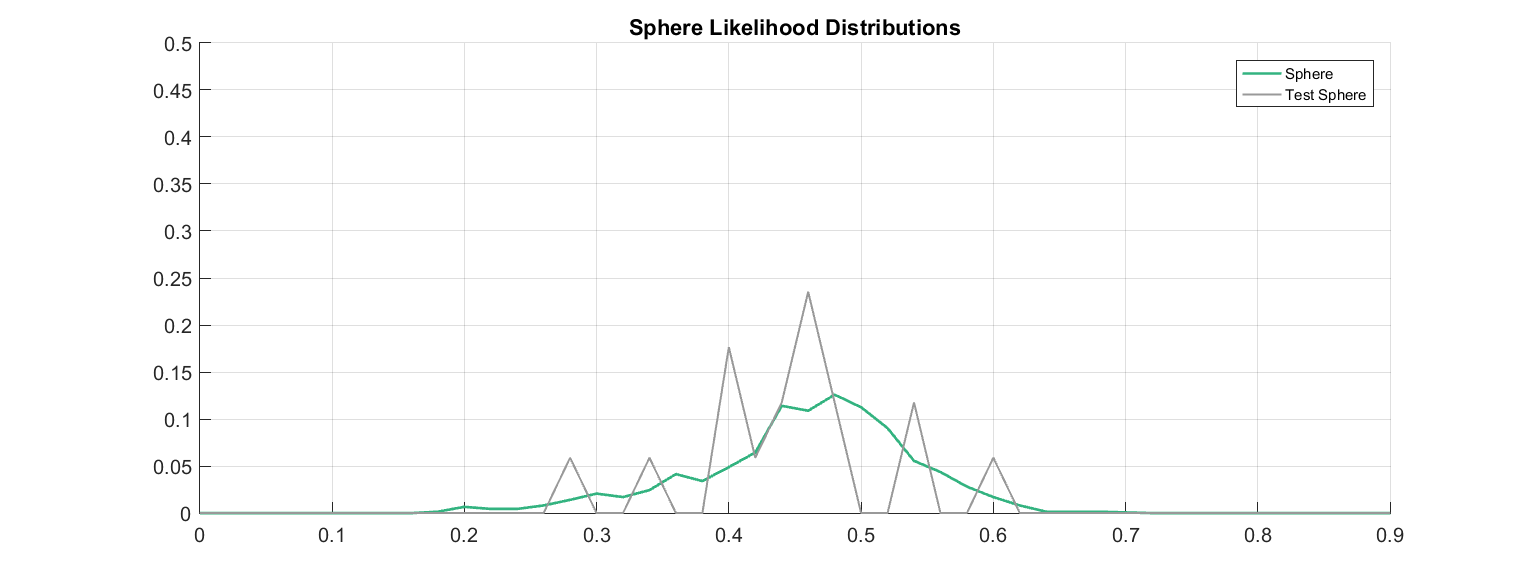}
\includegraphics[width=\columnwidth]{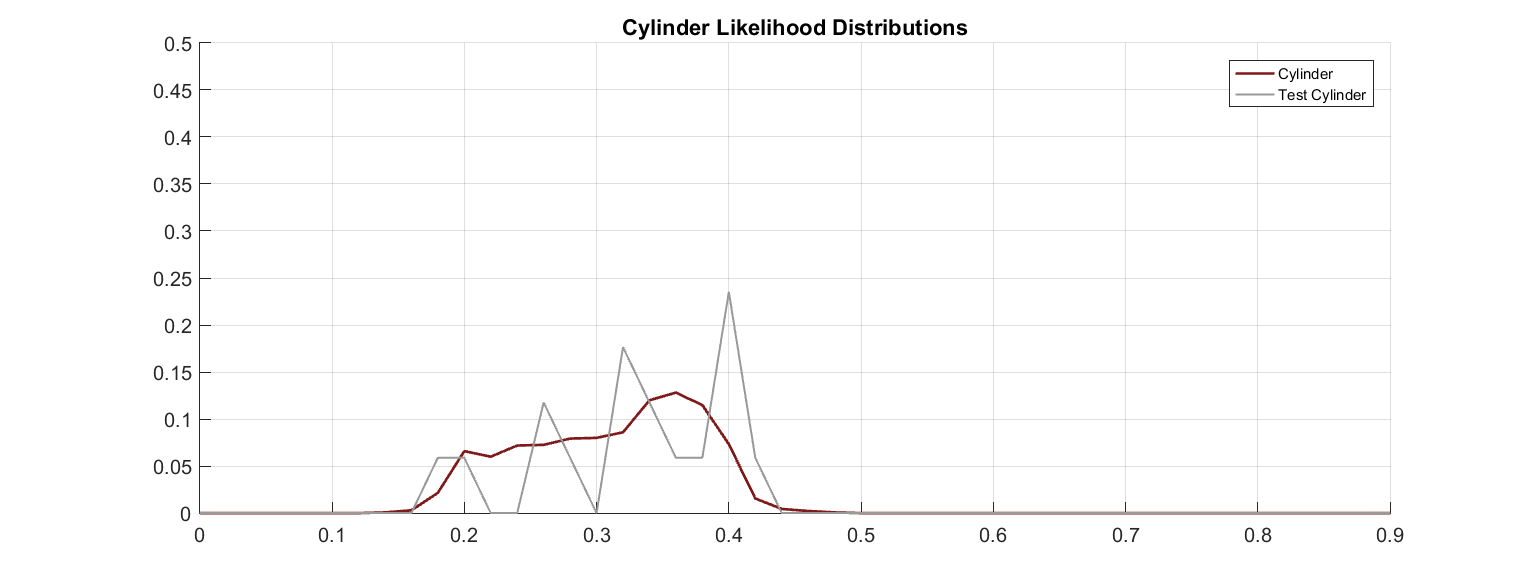}
\caption{Example anomaly distributions for test images coming from the training classes.  Each case did not earn a flag from their respective KS metrics.}\label{fig:notmalies}
\end{figure}

Of course, being able to distinguish between anomalies and reference distributions is only half the battle.  If our anomaly system cannot also hold back from flagging images that are \emph{not} foreign, then it is of quite limited usefulness.   To ensure this not the case, we took the same reference distributions and training images used for the torus anomaly detection and tested them against other non-foreign SAS images.  In all 12 cases, our anomaly framework succeeded in not flagging any of the images.  Figure \ref{fig:notmalies} shows an example for each class and, outside of some choppiness associated with the fewer number of samples, the general distributional trends matched.

Note that further tests using $\alpha=.0001$ also yielded the same results.  That is, because the two distributions were so different between the toruses and spheres, a ``safer'' confidence threshold could be utilized.  In a different experimental setting such a luxury may not be the case, but what we found here suggests that if an operator is looking to only flag anomalies that are truly unique to the training, there is some leeway in the choice of $\alpha$.

\section{Conclusion}\label{sec:discussion}

We have developed a robust Sonar ATR algorithm for real-world settings.  Our Bayesian spike and slab prior with class-specific parameters is key in allowing for the necessary discriminative power to differentiate between object classes.  In addition, the combination with discriminative dictionary learning concepts allows not only resiliency to noise as previous sparsity-based methods have done, but also an enhanced discrimination ability even in limited training regimes.  Further, the ability to identify targets foreign to the training set is a key component given the massive list of sea floor objects that could confuse even the most disciplined classifiers.  Flagging unknown targets can either save an operator from a costly extraction of what turns out to be a misshapen rock or alert this person to a uniquely shaped threatening device.

Our investigation into how well it does against multiplicative noise reveals a flexibility that PCS grants.  This algorithm can handle realistic bits of noise that are indeed seen in most sonar images and, even when an incredibly noisy image comes into play, the proposed PCS stands one of the best chances of correctly classifying such a nuisance. Future research may include further investigations into anomaly detection in conjunction with new class discovery methods for Sonar ATR based on sparsity based methods. 

\bibliographystyle{IEEEtran}
\bibliography{refTGARSS16}

\begin{thebibliography}{10}
\providecommand{\url}[1]{#1}
\csname url@samestyle\endcsname
\providecommand{\newblock}{\relax}
\providecommand{\bibinfo}[2]{#2}
\providecommand{\BIBentrySTDinterwordspacing}{\spaceskip=0pt\relax}
\providecommand{\BIBentryALTinterwordstretchfactor}{4}
\providecommand{\BIBentryALTinterwordspacing}{\spaceskip=\fontdimen2\font plus
\BIBentryALTinterwordstretchfactor\fontdimen3\font minus
  \fontdimen4\font\relax}
\providecommand{\BIBforeignlanguage}[2]{{%
\expandafter\ifx\csname l@#1\endcsname\relax
\typeout{** WARNING: IEEEtran.bst: No hyphenation pattern has been}%
\typeout{** loaded for the language `#1'. Using the pattern for}%
\typeout{** the default language instead.}%
\else
\language=\csname l@#1\endcsname
\fi
#2}}
\providecommand{\BIBdecl}{\relax}
\BIBdecl

\bibitem{hayes2009synthetic}
M.~P. Hayes and P.~T. Gough, ``Synthetic aperture sonar: a review of current
  status,'' \emph{Oceanic Engineering, IEEE Journal of}, vol.~34, no.~3, pp.
  207--224, 2009.

\bibitem{hansen2011introduction}
R.~E. Hansen, ``Introduction to synthetic aperture sonar,'' \emph{InTech},
  2011.

\bibitem{kriminger2015online}
E.~Kriminger, J.~Cobb, and J.~Principe, ``Online active learning for automatic
  target recognition,'' \emph{Oceanic Engineering, IEEE Journal of}, vol.~40,
  no.~3, pp. 583--591, July 2015.

\bibitem{stack2011automation}
J.~Stack, ``Automation for underwater mine recognition: current trends and
  future strategy,'' in \emph{SPIE Defense, Security, and Sensing}.\hskip 1em
  plus 0.5em minus 0.4em\relax International Society for Optics and Photonics,
  2011, pp. 80\,170K--80\,170K.

\bibitem{isaacs2015sonar}
J.~Isaacs, ``Sonar automatic target recognition for underwater uxo
  remediation,'' in \emph{Proceedings of the IEEE Conference on Computer Vision
  and Pattern Recognition Workshops}, 2015, pp. 134--140.

\bibitem{kumar2015robust}
N.~Kumar, U.~Mitra, and S.~S. Narayanan, ``Robust object classification in
  underwater sidescan sonar images by using reliability-aware fusion of shadow
  features,'' \emph{Oceanic Engineering, IEEE Journal of}, vol.~40, no.~3, pp.
  592--606, 2015.

\bibitem{cook2009analysis}
D.~A. Cook and D.~C. Brown, ``Analysis of phase error effects on stripmap
  sas,'' \emph{Oceanic Engineering, IEEE Journal of}, vol.~34, no.~3, pp.
  250--261, 2009.

\bibitem{wright2009robust}
J.~Wright, A.~Y. Yang, A.~Ganesh, S.~S. Sastry, and Y.~Ma, ``Robust face
  recognition via sparse representation,'' \emph{Pattern Analysis and Machine
  Intelligence, IEEE Transactions on}, vol.~31, no.~2, pp. 210--227, 2009.

\bibitem{vu2015dfdl}
T.~H. Vu, H.~S. Mousavi, V.~Monga, U.~Arvind~Rao, and G.~Rao, ``Dfdl:
  Discriminative feature-oriented dictionary learning for histopathological
  image classification,'' in \emph{Biomedical Imaging (ISBI), 2015 IEEE 12th
  International Symposium on}.\hskip 1em plus 0.5em minus 0.4em\relax IEEE,
  2015, pp. 990--994.

\bibitem{song2016sparse}
H.~Song, K.~Ji, Y.~Zhang, X.~Xing, and H.~Zou, ``Sparse representation-based
  sar image target classification on the 10-class mstar data set,''
  \emph{Applied Sciences}, vol.~6, no.~1, p.~26, 2016.

\bibitem{wagner2012toward}
A.~Wagner, J.~Wright, A.~Ganesh, Z.~Zhou, H.~Mobahi, and Y.~Ma, ``Toward a
  practical face recognition system: Robust alignment and illumination by
  sparse representation,'' \emph{Pattern Analysis and Machine Intelligence,
  IEEE Transactions on}, vol.~34, no.~2, pp. 372--386, 2012.

\bibitem{fandos2009sparse}
R.~Fandos, L.~Sadamori, and A.~Zoubir, ``Sparse representation based
  classification for mine hunting using synthetic aperture sonar,'' in
  \emph{Acoustics, Speech and Signal Processing (ICASSP), 2012 IEEE
  International Conference on}, March 2012, pp. 3393--3396.

\bibitem{kumar2012object}
N.~Kumar, Q.~F. Tan, and S.~S. Narayanan, ``Object classification in sidescan
  sonar images with sparse representation techniques,'' in \emph{Acoustics,
  Speech and Signal Processing (ICASSP), 2012 IEEE International Conference
  on}.\hskip 1em plus 0.5em minus 0.4em\relax IEEE, 2012, pp. 1333--1336.

\bibitem{mckay2015discriminative}
J.~McKay, R.~Raj, V.~Monga, and J.~Isaacs, ``Discriminative sparsity in sonar
  atr,'' \emph{Oceans 2015 Washington, DC}, 2015.

\bibitem{mckay2016localized}
J.~McKay, V.~Monga, and R.~Raj, ``Localized dictionary design for geometrically
  robust sonar atr,'' \emph{IGARSS 2016}, 2016.

\bibitem{srinivas2014simultaneous}
U.~Srinivas, H.~S. Mousavi, V.~Monga, A.~Hattel, and B.~Jayarao, ``Simultaneous
  sparsity model for histopathological image representation and
  classification,'' \emph{IEEE Transactions on Medical Imaging}, vol.~33,
  no.~5, pp. 1163--1179, May 2014.

\bibitem{mishne2013multiscale}
G.~Mishne and I.~Cohen, ``Multiscale anomaly detection using diffusion maps,''
  \emph{Selected Topics in Signal Processing, IEEE Journal of}, vol.~7, no.~1,
  pp. 111--123, 2013.

\bibitem{reed2003automatic}
S.~Reed, Y.~Petillot, and J.~Bell, ``An automatic approach to the detection and
  extraction of mine features in sidescan sonar,'' \emph{Oceanic Engineering,
  IEEE Journal of}, vol.~28, no.~1, pp. 90--105, 2003.

\bibitem{mignotte1999three}
M.~Mignotte, C.~Collet, P.~P{\'e}rez, and P.~Bouthemy, ``Three-class markovian
  segmentation of high-resolution sonar images,'' \emph{Computer Vision and
  Image Understanding}, vol.~76, no.~3, pp. 191--204, 1999.

\bibitem{goldman2005anomaly}
A.~Goldman and I.~Cohen, ``Anomaly subspace detection based on a multi-scale
  markov random field model,'' \emph{Signal Processing}, vol.~85, no.~3, pp.
  463--479, 2005.

\bibitem{wang2015object}
W.~Wang, B.~Cheng, and Y.~Chen, ``Object detection in side scan sonar,'' in
  \emph{Ninth International Symposium on Multispectral Image Processing and
  Pattern Recognition (MIPPR2015)}.\hskip 1em plus 0.5em minus 0.4em\relax
  International Society for Optics and Photonics, 2015, pp. 98\,110Z--98\,110Z.

\bibitem{mishne2015graph}
G.~Mishne, R.~Talmon, and I.~Cohen, ``Graph-based supervised automatic target
  detection,'' \emph{Geoscience and Remote Sensing, IEEE Transactions on},
  vol.~53, no.~5, pp. 2738--2754, 2015.

\bibitem{myers2010template}
V.~Myers and J.~Fawcett, ``A template matching procedure for automatic target
  recognition in synthetic aperture sonar imagery,'' \emph{Signal Processing
  Letters, IEEE}, vol.~17, no.~7, pp. 683--686, 2010.

\bibitem{groen2010model}
J.~Groen, E.~Coiras, J.~Del Rio~Vera, and B.~Evans, ``Model-based sea mine
  classification with synthetic aperture sonar,'' \emph{Radar, Sonar \&
  Navigation, IET}, vol.~4, no.~1, pp. 62--73, 2010.

\bibitem{fandos2011optimal}
R.~Fandos and A.~M. Zoubir, ``Optimal feature set for automatic detection and
  classification of underwater objects in sas images,'' \emph{Selected Topics
  in Signal Processing, IEEE Journal of}, vol.~5, no.~3, pp. 454--468, 2011.

\bibitem{isaacs2011laplace}
J.~C. Isaacs, ``Laplace-beltrami eigenfunction metrics and geodesic shape
  distance features for shape matching in synthetic aperture sonar,'' in
  \emph{Computer Vision and Pattern Recognition Workshops (CVPRW), 2011 IEEE
  Computer Society Conference on}.\hskip 1em plus 0.5em minus 0.4em\relax IEEE,
  2011, pp. 14--20.

\bibitem{williams2014exploiting}
D.~P. Williams and E.~Fakiris, ``Exploiting environmental information for
  improved underwater target classification in sonar imagery,'' \emph{IEEE
  Transactions on Geoscience and Remote Sensing}, vol.~52, no.~10, pp.
  6284--6297, Oct 2014.

\bibitem{fei2015contributions}
T.~Fei, D.~Kraus, and A.~Zoubir, ``Contributions to automatic target
  recognition systems for underwater mine classification,'' \emph{Geoscience
  and Remote Sensing, IEEE Transactions on}, vol.~53, no.~1, pp. 505--518, Jan
  2015.

\bibitem{dobeck1997automated}
G.~J. Dobeck, J.~C. Hyland \emph{et~al.}, ``Automated detection and
  classification of sea mines in sonar imagery,'' in \emph{AeroSense'97}.\hskip
  1em plus 0.5em minus 0.4em\relax International Society for Optics and
  Photonics, 1997, pp. 90--110.

\bibitem{isaacs2011signal}
J.~C. Isaacs and J.~D. Tucker, ``Signal diffusion features for automatic target
  recognition in synthetic aperture sonar,'' in \emph{Digital Signal Processing
  Workshop and IEEE Signal Processing Education Workshop (DSP/SPE), 2011
  IEEE}.\hskip 1em plus 0.5em minus 0.4em\relax IEEE, 2011, pp. 461--465.

\bibitem{azimi2000underwater}
M.~R. Azimi-Sadjadi, D.~Yao, Q.~Huang, and G.~J. Dobeck, ``Underwater target
  classification using wavelet packets and neural networks,'' \emph{Neural
  Networks, IEEE Transactions on}, vol.~11, no.~3, pp. 784--794, 2000.

\bibitem{candes2008introduction}
E.~J. Cand{\`e}s and M.~B. Wakin, ``An introduction to compressive sampling,''
  \emph{Signal Processing Magazine, IEEE}, vol.~25, no.~2, pp. 21--30, 2008.

\bibitem{ishwaran2005spike}
H.~Ishwaran and J.~S. Rao, ``Spike and slab variable selection: frequentist and
  bayesian strategies,'' \emph{Annals of Statistics}, pp. 730--773, 2005.

\bibitem{mousavi2015iterative}
H.~Mousavi, V.~Monga, and T.~Tran, ``Iterative convex refinement for sparse
  recovery,'' \emph{Signal Processing Letters, IEEE}, vol.~22, no.~11, pp.
  1903--1907, Nov 2015.

\bibitem{cevikalp2010face}
H.~Cevikalp and B.~Triggs, ``Face recognition based on image sets,'' in
  \emph{Computer Vision and Pattern Recognition (CVPR), 2010 IEEE Conference
  on}.\hskip 1em plus 0.5em minus 0.4em\relax IEEE, 2010, pp. 2567--2573.

\bibitem{kokiopoulou2010graph}
E.~Kokiopoulou and P.~Frossard, ``Graph-based classification of multiple
  observation sets,'' \emph{Pattern Recognition}, vol.~43, no.~12, pp.
  3988--3997, 2010.

\bibitem{fukui2005face}
K.~Fukui and O.~Yamaguchi, ``Face recognition using multi-viewpoint patterns
  for robot vision,'' in \emph{Robotics Research. The Eleventh International
  Symposium}.\hskip 1em plus 0.5em minus 0.4em\relax Springer, 2005, pp.
  192--201.

\bibitem{zhang2012pose}
N.~Zhang, R.~Farrell, and T.~Darrell, ``Pose pooling kernels for sub-category
  recognition,'' in \emph{Computer Vision and Pattern Recognition (CVPR), 2012
  IEEE Conference on}.\hskip 1em plus 0.5em minus 0.4em\relax IEEE, 2012, pp.
  3665--3672.

\bibitem{zhang2015survey}
Z.~Zhang, Y.~Xu, J.~Yang, X.~Li, and D.~Zhang, ``A survey of sparse
  representation: algorithms and applications,'' \emph{IEEE Access}, vol.~3,
  pp. 490--530, 2015.

\bibitem{srinivas2011sparsity}
U.~Srinivas, V.~Monga, Y.~Chen, and T.~D. Tran, ``Sparsity-based face
  recognition using discriminative graphical models,'' in \emph{2011 Conference
  Record of the Forty Fifth Asilomar Conference on Signals, Systems and
  Computers (ASILOMAR)}, Nov 2011, pp. 1204--1208.

\bibitem{kittler1998combining}
J.~Kittler, M.~Hatef, R.~P. Duin, and J.~Matas, ``On combining classifiers,''
  \emph{Pattern Analysis and Machine Intelligence, IEEE Transactions on},
  vol.~20, no.~3, pp. 226--239, 1998.

\bibitem{aharon2006ksvd}
M.~Aharon, M.~Elad, and A.~Bruckstein, ``K-svd: An algorithm for designing
  overcomplete dictionaries for sparse representation,'' \emph{Signal
  Processing, IEEE Transactions on}, vol.~54, no.~11, pp. 4311--4322, 2006.

\bibitem{chandola2009anomaly}
V.~Chandola, A.~Banerjee, and V.~Kumar, ``Anomaly detection: A survey,''
  \emph{ACM computing surveys (CSUR)}, vol.~41, no.~3, p.~15, 2009.

\bibitem{pimentel2014review}
M.~A. Pimentel, D.~A. Clifton, L.~Clifton, and L.~Tarassenko, ``A review of
  novelty detection,'' \emph{Signal Processing}, vol.~99, pp. 215--249, 2014.

\bibitem{darling1957kolmogorov}
D.~A. Darling, ``The kolmogorov-smirnov, cramer-von mises tests,'' \emph{The
  Annals of Mathematical Statistics}, vol.~28, no.~4, pp. 823--838, 1957.

\bibitem{kim2007interior}
S.-J. Kim, K.~Koh, M.~Lustig, S.~Boyd, and D.~Gorinevsky, ``An interior-point
  method for large-scale-regularized least squares,'' \emph{IEEE journal of
  selected topics in signal processing}, vol.~1, no.~4, pp. 606--617, 2007.

\bibitem{zhu2014model}
Z.~Zhu, X.~Xu, L.~Yang, H.~Yan, S.~Peng, and J.~Xu, ``A model-based sonar image
  atr method based on sift features,'' in \emph{OCEANS 2014 - TAIPEI}, April
  2014, pp. 1--4.

\bibitem{guo2015automatic}
Y.~Guo, Y.~Wang, D.~Kong, and X.~Shu, ``Automatic classification of
  intracardiac tumor and thrombi in echocardiography based on sparse
  representation,'' \emph{IEEE Journal of Biomedical and Health Informatics},
  vol.~19, no.~2, pp. 601--611, March 2015.

\end{thebibliography}

\end{document}